\pdfoutput=1

\documentclass[11pt]{article}

\usepackage[preprint]{acl}
\usepackage{pifont}
\usepackage{tabularx}
\usepackage{array}
\usepackage{booktabs}
\usepackage{hyperref}
\usepackage{multirow}
\usepackage{times}
\usepackage{latexsym}

\usepackage{colortbl}
\usepackage{xcolor}

\definecolor{bestcolor}{RGB}{144, 238, 144}  
\definecolor{secondbestcolor}{RGB}{204, 255, 204} 

\usepackage{colortbl}
\usepackage{pgfplots} 
\usepackage{geometry}
\definecolor{low}{RGB}{198,239,206}  
\definecolor{mid}{RGB}{255,255,153}  
\definecolor{high}{RGB}{255,199,206} 

\newcommand{\heatcell}[3]{%
  \pgfmathsetmacro{\percent}{(#1-#2)/(#3-#2)*100}
  \edef\tempcolor{\noexpand\cellcolor{low!\percent!high}}%
  \tempcolor #1
}

\usepackage[T1]{fontenc}

\usepackage[utf8]{inputenc}

\usepackage{microtype}

\usepackage{inconsolata}

\usepackage{graphicx}

\title{Follow the Flow: Fine-grained Flowchart Attribution \\ with Neurosymbolic Agents}

\author{Manan Suri $^ {\includegraphics[width=0.35cm]{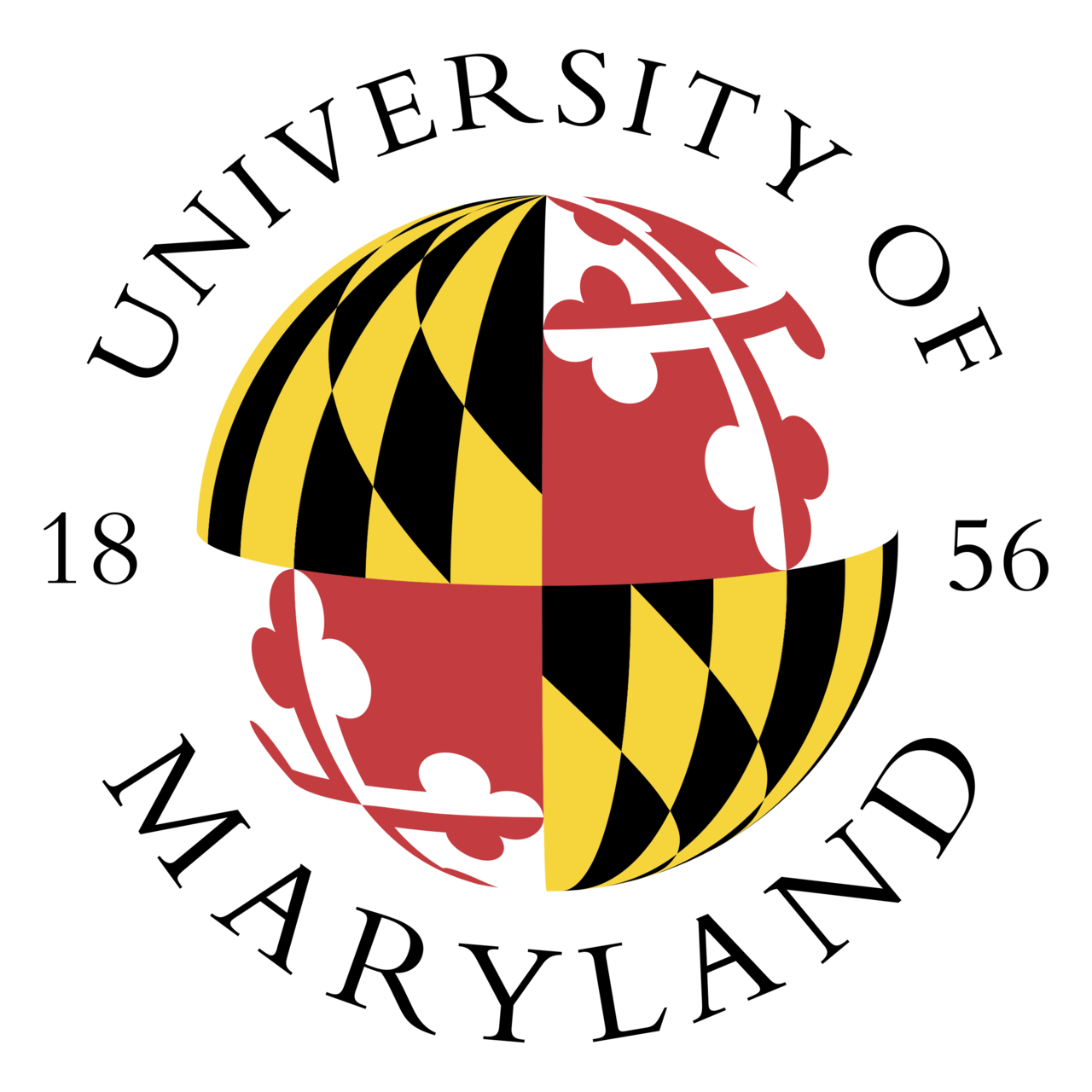}}$, \text{Puneet Mathur} $^ {{\includegraphics[width=0.2cm]{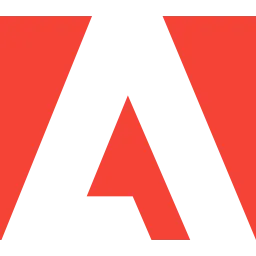}}}$ \thanks{Primary Research Mentor},  \textbf{\text{Nedim Lipka}} $^ {{\includegraphics[width=0.2cm]{figures/722666.png}} }$, \\ \textbf{\text{Franck Dernoncourt}}$ ^ {\includegraphics[width=0.2cm]{figures/722666.png}} $, \hspace{1.5pt} \textbf{\text{Ryan A. Rossi}} $^ {{\includegraphics[width=0.2cm]{figures/722666.png}} }$, \textbf{Vivek Gupta}$^ {\includegraphics[width=0.35cm]{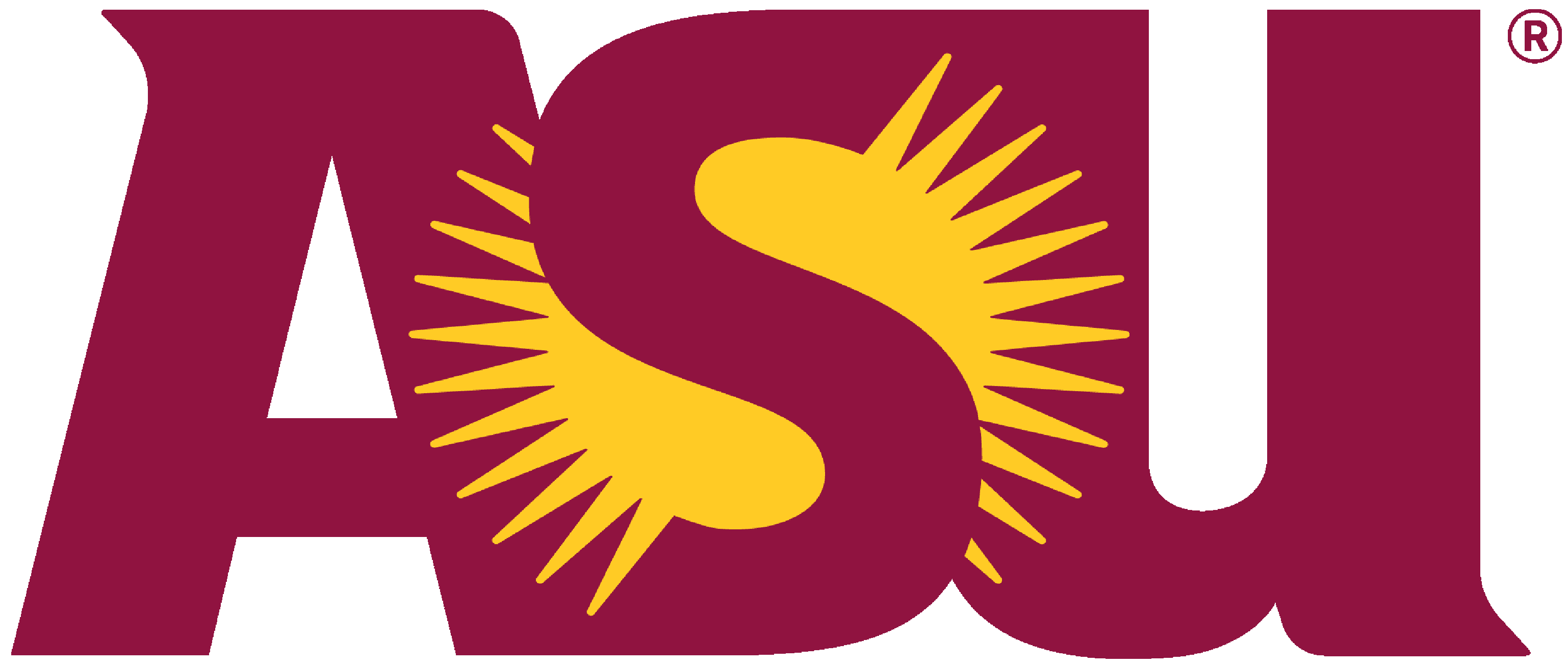}}$, \textbf{ \text{Dinesh Manocha}} $^ {\includegraphics[width=0.35cm]{figures/university-of-maryland-logo-1.png}}$\\ $^{\includegraphics[width=0.35cm]{figures/university-of-maryland-logo-1.png}}$ 
University of Maryland \hspace{10pt} $^{\includegraphics[width=0.2cm]{figures/722666.png}}$ Adobe Research \hspace{10pt} $^{\includegraphics[width=0.35cm]{figures/ASU-logo.png}}$ ASU   \\ \texttt{manans@umd.edu}, \texttt{puneetm@adobe.com}  }

\usepackage{amsmath}
\begin{document}
\maketitle
\begin{abstract}
Flowcharts are a critical tool for visualizing decision-making processes. However, their non-linear structure and complex visual-textual relationships make it challenging to interpret them using LLMs, as vision-language models frequently hallucinate nonexistent connections and decision paths when analyzing these diagrams. This leads to compromised reliability for automated flowchart processing in critical domains such as logistics, health, and engineering. We introduce the task of \texttt{Fine-grained Flowchart Attribution}, which traces specific components grounding a flowchart referring LLM response. Flowchart Attribution ensures the verifiability of LLM predictions and improves explainability by linking generated responses to the flowchart’s structure. We propose \texttt{FlowPathAgent}, a neurosymbolic agent that performs fine-grained post hoc attribution through graph-based reasoning. It first segments the flowchart, then converts it into a structured symbolic graph, and then employs an agentic approach to dynamically interact with the graph, to generate attribution paths. Additionally, we present \texttt{FlowExplainBench}, a novel benchmark for evaluating flowchart attributions across diverse styles, domains, and question types. Experimental results show that \texttt{FlowPathAgent} mitigates visual hallucinations in LLM answers over flowchart QA, outperforming strong baselines by 10-14\% on our proposed \texttt{FlowExplainBench} dataset.


\end{abstract}

\section{Introduction}
\begin{figure}[ht]
    \centering

    \includegraphics[width=\linewidth]{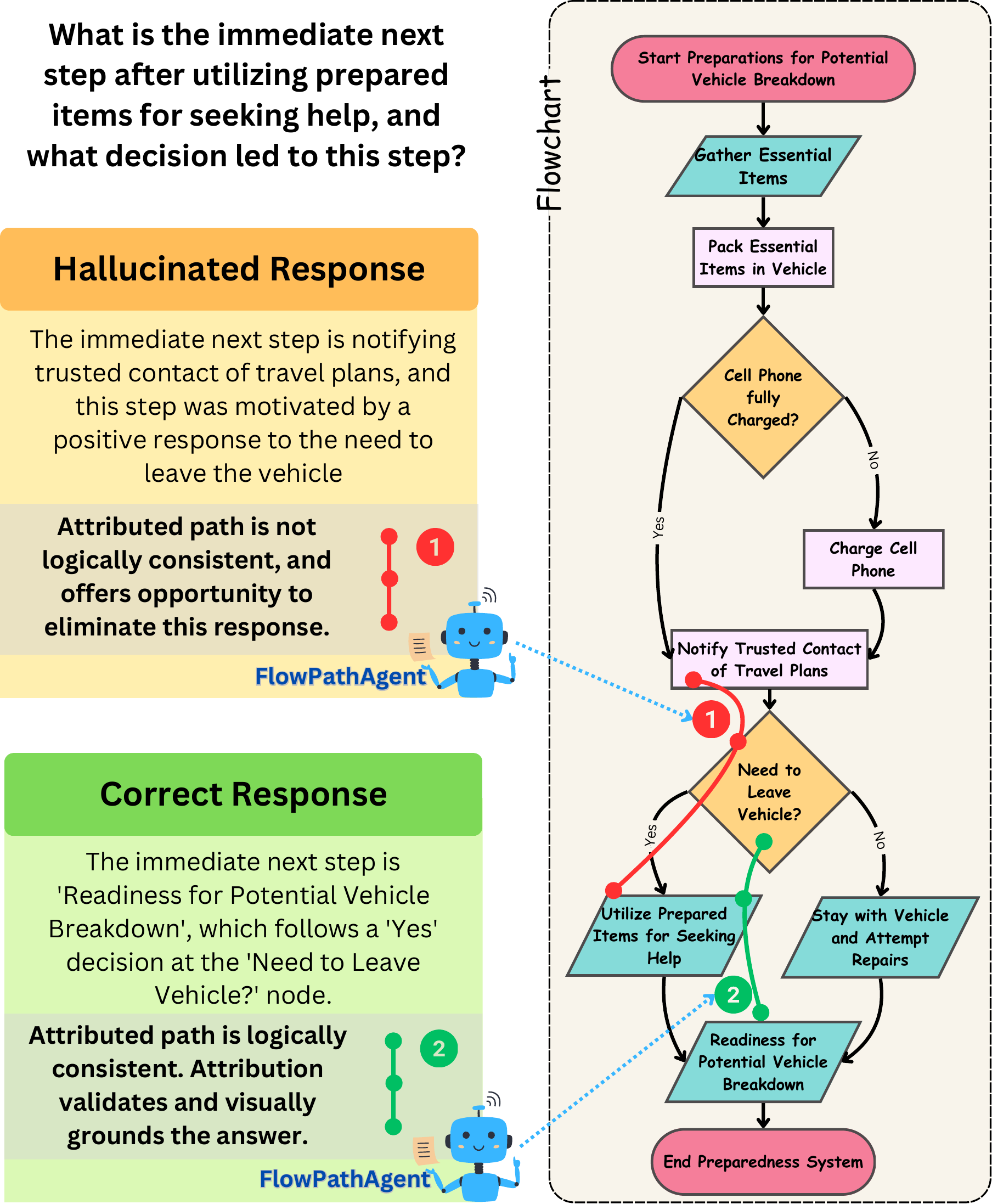}
    \caption{ \normalsize Attribution (represented by \tiny{$\bullet\!\!-\!\!\bullet\!\!-\!\!\bullet$}\normalsize) with \texttt{FlowPathAgent} ensures logical consistency in flowchart-based reasoning. \texttt{FlowPathAgent} uses a neurosymbolic approach to generate attribution paths  ( \textcolor[HTML]{ff3131}{\ding{202}} \&  \textcolor[HTML]{00bf63}{\ding{203}}) in the flowchart. This enhances interpretability and reliability in flowchart driven automated decision-making.}
    \label{fig:intro}
\end{figure}

Flowcharts are a fundamental tool for representing structured decision-making processes. Used across domains such as software engineering, business process modeling, and instructional design, flowcharts provide a visual roadmap of logical operations, guiding both human users and automated systems \cite{flowchart_code, flowchart_finance, flowchart_education, flowchart_multiple}. Their structured yet visual nature makes them an effective medium for conveying procedural logic. However, interpreting flowcharts accurately is challenging due to their nonlinear structures (branching and loop-based control flow), where meaning emerges from the interplay between textual content, visual arrangement, and logical dependencies. Ambiguities in flowchart interpretation arise from diverse notational conventions, implicit relationships, and misinferred steps, making precise attribution of information sources difficult \cite{ambuguity}.


Recent advancements in Vision Language Models (VLMs) have enabled substantial progress in flowchart processing \cite{flowvqa}. These models leverage both textual and visual information, allowing them to extract structural relationships, recognize decision nodes, and generate answers based on flowchart content. However, despite their capabilities, VLMs struggle with hallucination: the tendency to generate information that is not grounded in the input \cite{huang2024visual, guan2024hallusionbench}. In the context of flowcharts, hallucination can manifest as misidentifying decision nodes, producing incorrect logical pathways, or fabricating connections that do not exist in the original structure. This issue severely impacts the reliability of automated flowchart reasoning, particularly in high-stakes applications such as healthcare, software verification and process automation.

Although VLMs have made significant progress in understanding flowcharts, prior work has mainly concentrated on flowchart parsing \cite{arbaz2024genflowchartparsing}, conversion \cite{shukla2023towards, liu-etal-2022-code}, and question-answering \cite{flowvqa, tannert2022first}, while overlooking the critical aspect of fine-grained attribution. While existing attribution methods \cite{huo2023retrieving, chen2023complex} focus on textual grounding, attributing responses to visual-textual elements like flowcharts presents unique challenges. It involves not just text recognition, but also interpreting the interconnected decision nodes, hierarchical structures, and conditional pathways that define flowchart semantics. Attribution serves as a crucial mechanism for mitigating hallucination by explicitly tracing the paths in the flowchart that ground a particular response, enabling rigorous evaluation of the model's fidelity to the flowchart's logic, as illustrated in Fig \ref{fig:intro}. Such fine-grained attribution is fundamental for ensuring reliability, particularly when these systems are deployed in domains where verifiable decision-making is crucial.



\noindent\textbf{Main Results.} We introduce Flowchart Attribution task aimed at identifying the optimal path within a flowchart that grounds the model’s response. The optimal path aims to extract the most relevant sequence of nodes and edges that directly support the model’s reasoning, encompassing all the key decision points and actions involved in the prediction. To facilitate the evaluation of this task, we propose \texttt{FlowExplainBench}, a novel benchmark that features a diverse set of flowcharts with varying styles, domains, and question types.

We introduce \texttt{FlowPathAgent}, a neurosymbolic agent specifically designed to perform fine-grained as a post-hoc flowchart attribution. Instead of relying solely on text-based or vision-based cues, \texttt{FlowPathAgent} integrates symbolic reasoning by using an agentic interface to interact with the flowchart as a graph object. \texttt{FlowPathAgent} begins with segmenting flowcharts into distinct components, followed by constructing symbolically operable flowchart representations. These graph-based representations have direct correspondence to visual regions of the flowchart, enabling the model to interoperate between the visual and symbolic representations. We leverage graph tools to extract and manipulate these representations, allowing for identification of relevant nodes and edges. Our methodology facilitates precise attribution of the model’s reasoning steps to specific decision points within the flowchart, providing accurate and interpretable explanations of the model’s output. Experimental results demonstrate that \texttt{FlowPathAgent} significantly outperforms strong baselines \cite{lai2024lisa, peng2023kosmos, yuan2025sa2vamarryingsam2llava} by 10-14\% on \texttt{FlowExplainBench}. 

\noindent Our \textbf{main contributions}\footnote{\href{https://anonymous.4open.science/r/flowpathagent-8E15}{Code and data will be released on acceptance.}} are:

\begin{itemize}
    \item We introduce a new task of \textbf{Fine-grained Flowchart Attribution} where the goal is to identify the optimal path within a flowchart diagram that grounds the LLM text response.
    \item \textbf{\texttt{FlowExplainBench}} - a novel evaluation benchmark consisting of 1k+ high quality attribution annotations over flowchart QA with diverse styles, domains, and question types.
    \item \textbf{\texttt{FlowPathAgent}} - a neurosymbolic agent capable of performing fine-grained post-hoc attribution for flowchart QA. \texttt{FlowPathAgent} uses a VLM-based agentic approach to perform graph-based reasoning and symbolic manipulation to accurately trace the decision process within flowcharts.
\end{itemize}

\section{Related Work}
\subsection{Flowchart Understanding}
Research in flowchart understanding has evolved from basic image processing to complex reasoning tasks. Modern deep learning approaches, such as FR-DETR \cite{frdetr}, have significantly improved symbol and edge detection through end-to-end architectures that combine CNN backbones with multi-scale transformers. The emergence of LLMs has led to  benchmarks like FlowchartQA \cite{tannert2022first}, FlowLearn \cite{pan2024flowlearn}, SCI-CQA \cite{shen2024rethinking}, and FlowVQA \cite{flowvqa}, which assess geometric understanding, spatial reasoning, and logical capabilities of models for question-answering on flowchart images . Recent work like \cite{ye2024endtoendvlmsleveragingintermediate} has begun exploring alternatives to end-to-end VLMs; \cite{ye2024endtoendvlmsleveragingintermediate} introduced intermediate textual representations between visual processing and reasoning steps for Flowchart QA; \cite{liu-etal-2022-code, shukla2023towards} explored code generation from flowcharts. 

\subsection{Attribution in LLMs}
Large Language Models (LLMs) are challenged with factual accuracy \cite{hallu1}. While various solutions have emerged, including citation-aware training \cite{gao2023enabling} and tool augmentation \cite{ye2023effective}, ensuring reliable attributions remains crucial. Three primary attribution strategies have emerged in literature: (1) Direct model-driven attribution generates answers and attributions simultaneously \cite{peskoff2023credible, sun2022recitation}. (2) Post-retrieval answering retrieves information before answering \cite{ye2023effective, li2023llatrieval, huo2023retrieving, chen2023complex}. (3) Post-hoc attribution generates answers first and then searches for supporting references \cite{li2023survey}. Our work falls in the scope of Post-hoc attribution, as it serves as a modular approach integrable with existing system, without accessing the response generation mechanism. Recent work has expanded attribution capabilities to handle diverse data formats. While MATSA \cite{mathur2024matsa} explored fine-grained attribution for tables through a multi-agent approach, VISA \cite{ma2024visaretrievalaugmentedgeneration} advanced visual attribution by leveraging vision-language models to highlight specific regions in document screenshots. Ours is the frst work on flowchart QA attribution.

\section{Post-hoc Flowchart Attribution}
\label{sec:task}

We formalize fine-grained post-hoc Flowchart Attribution as  follows:
Given a dataset \( \mathcal{D} \) consisting of a set of flowchart images \( \mathcal{F} \), each flowchart image \( c_i \in \mathcal{F}, c_i = \mathcal{I}^{w \times h \times 3}\) corresponds to a logical graph representation \( G_i = (V_i, E_i) \), where \( V_i \) represents the set of nodes and \( E_i \) represents the edges between them. Each node corresponds to a logical operation or directive statement, and the edges represent the flow between these operations. Additionally, the input includes a flowchart-referring statement \( s_i \), which is a natural language description of a process or action to be grounded in the flowchart image. The underlying goal is to find a path in the image that grounds the statement \( s_i \). This path may be disjoint, but it should correspond to a set of regions in the flowchart image. The regions are the physical abstraction that corresponds to the logical nodes in the graph. Formally, the task can be represented as a mapping function:

\[
F: (c_i, s_i) \mapsto \mathcal{R}_{s_i},
\]

\noindent where \( F \) maps the flowchart image \( c_i \) and the statement \( s_i \) to a set of regions \( \mathcal{R}_{s_i} \) in the image. \( \mathcal{R}_{s_i} = \{r_{i1}, r_{i2}, \dots, r_{in}\} \) represents the sequence of regions in the image that correspond to a path of logical nodes, and the edges included between consecutive nodes {\( v_{i1}, v_{i2}, \dots, v_{in} \)} in the graph \( G_i \), grounding the statement \( s_i \). The path may be disjoint, but it should satisfy the following criteria:

\noindent\textbf{1. Optimality}: The path should be the shortest sequence of regions that ground the statement \( s \).

\noindent\textbf{2. Contextual Alignment}: The path should correspond to the relevant actions and decisions 
described in \( s \), matching the flow of the process.

\noindent\textbf{3. Exclusivity}: No additional regions outside of \( \mathcal{R}_{s_i}\) are necessary to fully explain the statement \( s \).

\section{\texttt{FlowExplainBench}}

To enable systematic evaluation of flowchart attribution, we introduce \texttt{FlowExplainBench}, a comprehensive benchmark designed with four key criteria: diverse visual styles, varied question types, multiple flowchart domains, and faithful ground-truth attributions (see Table \ref{tab:dataset}). Each entry in the dataset consists of the following components: the flowchart image \( c \), a statement \( s \) (which, in this context, is a Question-Answer pair), a set of attributed logical nodes \( v_1, v_2, \dots, v_n \), and their corresponding visual regions \( \mathcal{R}_s = \{r_1, r_2, \dots, r_n\} \). These visual regions represent the physical abstractions of the logical nodes, which are mapped from the flowchart image \( c \) as discussed in section \ref{sec:task}.


\subsection{Data Sources}

\texttt{FlowExplainBench} is constructed using the test split of the FlowVQA dataset \cite{flowvqa}. This dataset comprises high-quality flowchart images sourced from diverse domains, including the FloCo dataset, which emphasizes code-related flowcharts \cite{shukla2023towards}, as well as widely recognized DIY platforms such as Wikihow and Instructables. These sources contribute to three distinct data splits: \textit{Code}, \textit{Wiki}, and \textit{Instruct}. For each flowchart, corresponding Mermaid code and metadata (e.g., original code and process summaries) are included. The dataset contains four question types: Fact retrieval, Applied Scenario, Flow Referential, and Topological.


\begin{figure*}[t]
    \centering
    \includegraphics[width=\linewidth]{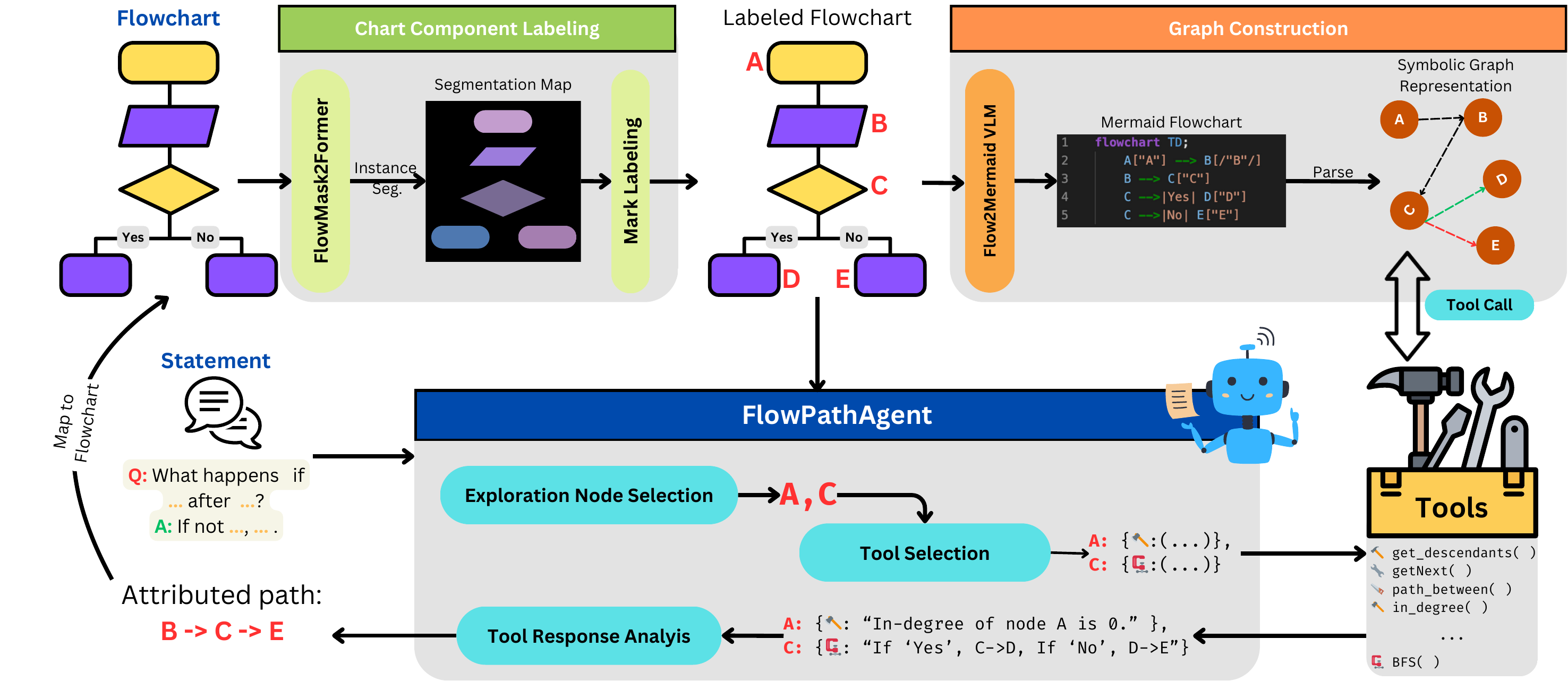}
    \caption{Overview of \textbf{\texttt{FlowPathAgent}}. \texttt{FlowPathAgent} processes a flowchart image through segmentation-based component labeling, constructs a symbolic graph representation using Mermaid, and employs a neurosymbolic agent, that treats the flowchart as a symbolic graph to attribute nodes based on an input statement. The agent interacts with predefined tools to analyze and traverse the flowchart structure, producing attributions as interpretable mappings of relevant nodes back onto the original flowchart.}
    \label{fig:flowpathagent}
\end{figure*}

\begin{table}[]
\resizebox{\columnwidth}{!}{%
\begin{tabular}{l|r|r|r|r|}
\cline{2-5}
 & \multicolumn{1}{l|}{\textbf{Code}} & \multicolumn{1}{l|}{\textbf{Wiki}} & \multicolumn{1}{l|}{\textbf{Instruct}} & \multicolumn{1}{l|}{\textbf{Overall}} \\ \hline
\multicolumn{1}{|l|}{\textbf{\# of Flowcharts}} & 189 & 470 & 294 & 953 \\
\multicolumn{1}{|l|}{\textbf{\# of Questions}} & 246 & 610 & 382 & 1238 \\
\multicolumn{1}{|l|}{\hspace{10pt}Fact Retrieval} & 88 & 163 & 102 & 353 \\
\multicolumn{1}{|l|}{\hspace{10pt}Applied Scenario} & 69 & 128 & 90 & 287 \\
\multicolumn{1}{|l|}{\hspace{10pt}Flow Referential} & 43 & 128 & 87 & 258 \\
\multicolumn{1}{|l|}{\hspace{10pt}Topological} & 46 & 191 & 103 & 340 \\
\multicolumn{1}{|l|}{\textbf{Avg \# of Nodes}} & 11.85 & 24.49 & 21.59 & 21.08 \\
\multicolumn{1}{|l|}{\textbf{Max \# of Nodes}} & 29 & 43 & 44 & 44 \\
\multicolumn{1}{|l|}{\textbf{Avg Attributed Path Length}} & 2.59 & 3.21 & 2.88 & 2.99 \\
\multicolumn{1}{|l|}{\textbf{Max Attributed Path Length}} & 15 & 35 & 21 & 35 \\
\multicolumn{1}{|l|}{\textbf{Avg Words (Question)}} & 26.99 & 26.12 & 26.56 & 26.43 \\
\multicolumn{1}{|l|}{\textbf{Avg Words (Answer)}} & 8.62 & 8.74 & 9.50 & 8.95 \\ \hline
\end{tabular}%
}
\caption{Detailed overview of distribution and characteristics 
 of constituent splits of \texttt{FlowExplainBench}.}
\label{tab:dataset}
\end{table}
\subsection{Visual Diversity}

To ensure that \texttt{FlowExplainBench} represents a broader spectrum of flowchart styles encountered in real-world applications, we introduce four distinct style types for flowchart generation: \textbf{1. Single Color:} Flowcharts that use a single color for all nodes throughout the chart for simplicity and visual cohesion. \textbf{2. Multi Color:} Flowcharts that utilize multiple colors, sampled from a palette to represent different nodes. \textbf{3. Default Mermaid:} The standard Mermaid styling as found in FlowVQA. \textbf{4. Black and White:} Flowcharts designed using only black and white elements. For the \textbf{Single Color} style, we incorporate 40 unique colors, while for the \textbf{Multi Color} style, we use 35 curated color palettes each containing 4 to 5 colors. We first generate SVG flowcharts from source Mermaid code, subsequently injecting templated CSS into the SVGs to implement the desired styles. Finally, the SVGs are converted into PNG images, and the regions of interest (i.e., the flowchart nodes) are defined using the positional and shape information derived from the SVG metadata.

\subsection{Attribution Annotation}

The attribution annotation process is as follows:

\noindent\textbf{Step 1: Automatic Labeling.}  
We use GPT-4 to perform the initial attribution for corresponding QA pairs directly in the Mermaid source code. By analyzing the nature of different question types, we generalize the attribution patterns and provide GPT-4 with few-shot examples in the prompt.

\noindent\textbf{Step 2: Human Verification.}  
Two human evaluators are involved in the next stage, where they interact with an attribution platform that allows them to select nodes in the flowchart to be attributed. The inter-annotator agreement, measured using Cohen’s Kappa ($\kappa$), shows a high level of agreement both between the two annotators ($\kappa$ = 0.89) and between the annotators and the initial GPT-4-generated labels ($\kappa$ = 0.72, 0.80). More details on human annotation in Appendix Sec. \ref{annotation}.

\noindent\textbf{Step 3: Multi-step Question Filtering}. We applied a filtering srategy to get rid of trivial and low-quality QA pairs. Questions related to node and edge count were excluded, as they required trivially attributing the entire graph rather than reasoning over its fine-grained individual components. This excluded 1792 samples from the annotation exercise described above. Subsequently, for each flowchart image, questions with the highest agreement among annotators were selected, prioritizing cases where both human annotators concurred. This yielded an initial set of 953 samples. To achieve balance across three domains and four question types, additional high-agreement samples were selected from underrepresented categories, resulting in a final benchmark of 1,238 samples.

\section{\texttt{FlowPathAgent}}
\texttt{FlowPathAgent} (Fig \ref{fig:flowpathagent}), is a neurosymbolic agent designed for structured reasoning over flowcharts for fine-grained flowchart attribution. The approach consists of three key stages: \textbf{Chart Component Labeling}, which segments and labels chart components; \textbf{Graph Construction}, which constructs a symbolic graph from the labeled flowchart; and \textbf{Neurosymbolic Agent-based Analysis}, which uses graph-based tools to interact with the symbolic flowchart to generate attributed paths. Each stage plays a critical role in bridging the gap between visual representations and symbolic reasoning over structured workflows.

\subsection{Chart Component Labeling}
We identify and label individual flowchart components, ensuring an explicit correspondence between visual elements and the symbolic representations generated in subsequent steps. 

\noindent\textbf{FlowMask2Former.} To achieve flowchart component recognition, we construct a synthetic dataset using the training split of FlowVQA \cite{flowvqa}, incorporating style diversification techniques similar to those described in Section 4.2, but with different color schemes. Further the node content is replaced with randomized text. These augmentations improve domain generalization and ensure robust performance across diverse flowchart styles. We fine-tune Mask2Former \cite{mask2aformer} on this dataset for instance segmentation, specifically targeting node recognition. The fine-tuned model, FlowMask2Transformer, generates segmentation maps, from which individual nodes are sequentially labeled using alphabetical identifiers, rendered in red text on the flowchart image, to serve as visual anchors for graph construction, reasoning, and node referencing.

\subsection{Graph Construction}
Flowcharts inherently encode structured logical processes, making graph-based representations ideal for symbolic reasoning. By converting visual flowcharts into symbolic graph structures, we eliminate reliance on visual recognition for every reasoning step, ensuring robust handling of distant relationships that visual models often misinterpret. The symbolic graph facilitates efficient graph-based operations such as traversal, topological analysis, and conditional evaluation. This structured representation also enhances interpretability and enables automated verification of logical consistency. Moreover, as flowchart complexity increases, our method avoids the compounding errors seen in purely visual models by explicitly encoding edge conditions and node relationships, enabling reliable and scalable path tracing.

\noindent\textbf{Flow2Mermaid VLM.} To convert the labeled flowchart to a symbolic graph representation, we first convert the visual flowchart to a Mermaid code, and then parse the Mermaid code to generate the symbolic graph.
For the Flowchart to Mermaid transformation, we fine-tune Qwen2-VL(7B) \cite{wang2024qwen2} using supervised fine-tuning (SFT) on a style-diversified projection of the FlowVQA \cite{flowvqa} training set, with marked alphabetic node labels sourced from SVG metadata. Flow2Mermaid VLM  is trained to generate Mermaid flowchart code directly from flowchart images, using the alphabetical node labels as anchors to maintain consistency between visual and symbolic representations. We perform fine-tuning to improve the ability to generate accurate and semantically robust Mermaid syntax, minimize structural inconsistencies that could affect graph analyses, and adapt to the varied aspect ratios and visual styles found in flowcharts. 


The generated Mermaid representation is then parsed into a symbolic graph, tailored to capture the specific properties of flowcharts, including boolean conditional edges and node-level statement mappings. Additionally, we define a comprehensive suite of tools to operate on this symbolic graph, enabling structured function calls for reasoning over the flowchart’s logical structure. The list of tools and their API is described in the Appendix.

\subsection{Neurosymbolic Agent}
\texttt{FlowPathAgent} employs a neurosymbolic reasoning approach to attribute relevant nodes based on an input statement. In this context, it combines neural models i.e. VLMs to plan, reason and attribute in a discrete token space, based on observations made via tool use over a symbolic graph representing the flowchart. The agent operates on a sequence of interdependent steps:

\noindent\textbf{1. Node Selection:} During the initial planning stage, our agent identifies nodes to be explored by referencing their corresponding labels in the flowchart image. Additionally, it clarifies the expectation and underlying rationale for each node selection. This is the only step where the labeled flowchart image is passed to the underlying VLM.

\noindent\textbf{2. Tool Selection:} Our agent employs reasoning-based prompting to determine the necessary symbolic tools and their respective functional parameters for the selected nodes.

\noindent\textbf{3. Tool Execution:} The selected tools are executed on the symbolic graph representation to extract relevant insights. Multiple sequential cycles of Tool Selection and Tool Execution may occur, with each cycle selecting and executing a single tool.

\noindent\textbf{4. Tool Response Analysis:} The agent interprets observations from tool-use, in relation to the given statement, generating a path of nodes in the flowchart that attribute the statement.

\noindent\textbf{5. Mapping to Original Flowchart:} Finally, the attributed path's node labels are mapped back onto the flowchart image using the segmentation regions obtained during the labeling stage.

\section{Experimental Set-up}
\subsection{Baselines}

\textbf{Zero-shot GPT-4o} \cite{openai_gpt4o_2024} predicts normalized bounding box coordinates for zero-shot localization \cite{yang2023dawn}.

\noindent\textbf{Kosmos-2:}\cite{peng2023kosmos} performs referring expression grounding and bounding box generation by linking objects in images with text.

\begin{table*}[]
\centering
\resizebox{2\columnwidth}{!}{%
\begin{tabular}{l|ccc|ccc|ccc|ccc}
\hline
\multirow{2}{*}{\textbf{Baseline}} & \multicolumn{3}{c|}{\textbf{Overall}} & \multicolumn{3}{c|}{\textbf{FEBench-Code}} & \multicolumn{3}{c|}{\textbf{FEBench-Wiki}} & \multicolumn{3}{c}{\textbf{FEBench-Instruct}} \\
 & \textbf{Precision} & \textbf{Recall} & \textbf{F1} & \textbf{Precision} & \textbf{Recall} & \textbf{F1} & \textbf{Precision} & \textbf{Recall} & \textbf{F1} & \textbf{Precision} & \textbf{Recall} & \textbf{F1} \\ \hline
Kosmos-2 \cite{peng2023kosmos} & 37.14 & 1.76 & 3.36 & 41.41 & 6.45 & 11.16 & 20.69 & 0.31 & 0.60 & 38.30 & 1.64 & 3.14 \\
LISA \cite{lai2024lisa} & 18.01 & 14.34 & 15.97 & 35.36 & 19.18 & 24.87 & 14.09 & 11.74 & 12.81 & 18.45 & 16.18 & 17.24 \\
SA2VA \cite{yuan2025sa2vamarryingsam2llava} & 66.36 & 9.88 & 17.20 & 79.35 & 19.34 & 31.10 & 58.47 & 7.40 & 13.14 & 65.99 & 8.82 & 15.56 \\
VisProg \cite{Gupta2022VisProg} &45.95 & 0.46 & 0.91 & 46.88& 2.30 & 4.49  & 0.00 & 0.00 & 0.00 &  25.00 & 00.09 & 0.18  \\
GPT4o Bounding Box & 58.82 & 1.90 & 3.68 & \cellcolor{bestcolor}80.00 & 1.89 & 3.69 & 53.19 & 1.29 & 2.51 & 57.89 & 3.00 & 5.70 \\
GPT4o SoM & \cellcolor{secondbestcolor}74.10 & \cellcolor{secondbestcolor}67.69 & \cellcolor{secondbestcolor}70.75 & 67.32 & \cellcolor{secondbestcolor}70.28 & \cellcolor{secondbestcolor}68.77 & \cellcolor{secondbestcolor}74.55 & \cellcolor{secondbestcolor}65.03 & \cellcolor{secondbestcolor}69.47 & \cellcolor{secondbestcolor}77.84 & \cellcolor{secondbestcolor}70.91 & \cellcolor{secondbestcolor}74.22 \\ \hline
\textbf{\texttt{FlowPathAgent}} & \cellcolor{bestcolor}77.19 & \cellcolor{bestcolor}77.21 & \cellcolor{bestcolor}77.20 & \cellcolor{secondbestcolor}74.18 & \cellcolor{bestcolor}80.62 & \cellcolor{bestcolor}77.27 & \cellcolor{bestcolor}76.29 & \cellcolor{bestcolor}74.21 & \cellcolor{bestcolor}75.23 & \cellcolor{bestcolor}80.28 & \cellcolor{bestcolor}80.19 & \cellcolor{bestcolor}80.23 \\ \hline
\end{tabular}%
}
\caption{Performance comparison of \texttt{FlowPathAgent} with baselines on \texttt{FlowExplainBench}. \colorbox{bestcolor}{Best} and \colorbox{secondbestcolor}{second-best} results have been highlighted.}

\label{tab:results}
\end{table*}

\noindent\textbf{LISA:} \cite{li2023llatrieval} generates segmentation masks from textual queries with minimal fine-tuning on task-specific data.

\noindent\textbf{SA2VA}\cite{yuan2025sa2vamarryingsam2llava} combines SAM-2 and LLaVA for referring segmentation.

\noindent\textbf{VisProg} \cite{Gupta2022VisProg} agent generates visual programs by decomposing queries into executable steps for explainable visual reasoning.

\noindent\textbf{GPT4o + FlowMask2Former SoM Prompting} ablation uses GPT-4o on segmented flowchart generated by FlowMask2Former and applies Set-of-Marks (SoM) \cite{yang2023set}. More details on baselines in Appendix Sec.\ref{baselines}.

\subsection{Evaluation}
To map the segmented regions to the ground truth nodes, we apply an Intersection over Union (IoU) threshold of 0.7 to ensure high fidelity between ground-truth and predicted nodes. The ground-truth node with the maximum overlap is selected as the reference for the segmented node. This process is crucial for fine-grained attribution, where accurate identification of individual flowchart components is required. For each baseline, we collect the nodes identified by the model and treat the set of ordered nodes as the attributed path. We then compute the micro-averaged Precision, Recall, and F1 scores to assess the model's performance.

\noindent More extensive experimental details have been provided in Appendix Sec. \ref{implementation_details}.




\section{Results and Discussion}

\textbf{Baseline Comparison.}
\texttt{FlowPathAgent} demonstrates a significant improvement over all baseline models when evaluated on the \texttt{FlowExplainBench}, outperforming them by a margin of 6-65 percentage points, as shown in Table \ref{tab:results}. Visual grounding models, including Kosmos-2, LISA, and SA2VA, exhibit suboptimal performance. This is primarily due to their limited ability to process visual logic, which is crucial for fine-grained flowchart attribution. These models struggle to correctly map logical relationships between elements in the flowchart, resulting in less accurate attributions.

\begin{figure}[ht]
    \centering
    \includegraphics[width=\linewidth]{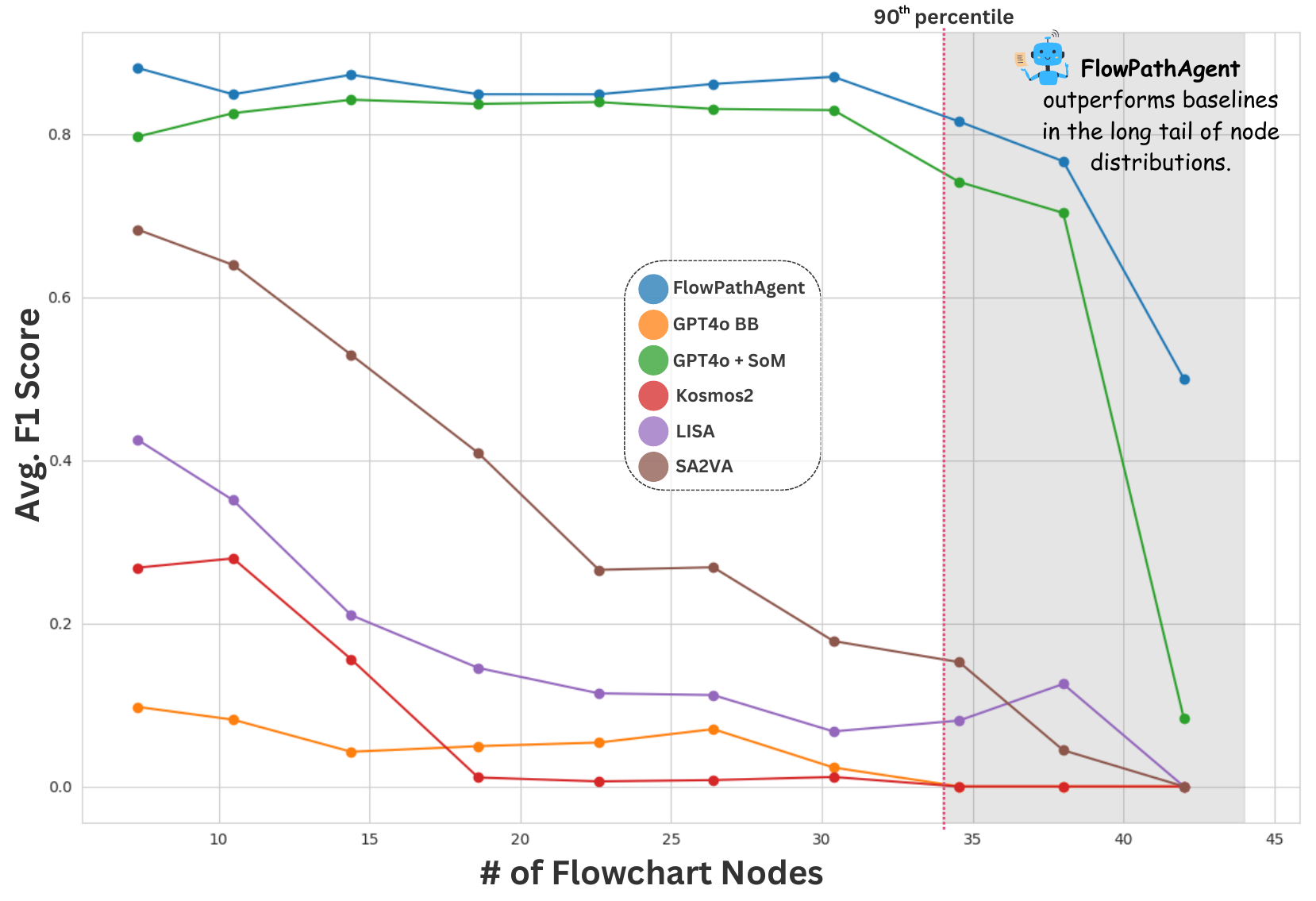}
    \caption{Performance comparison of \texttt{FlowPathAgent} against baselines demonstrates superior effectiveness across long-tail distribution of node count in flowcharts.}
    \label{fig:longtail}
\end{figure}

\begin{figure}
    \centering
    \includegraphics[width=0.8\linewidth]{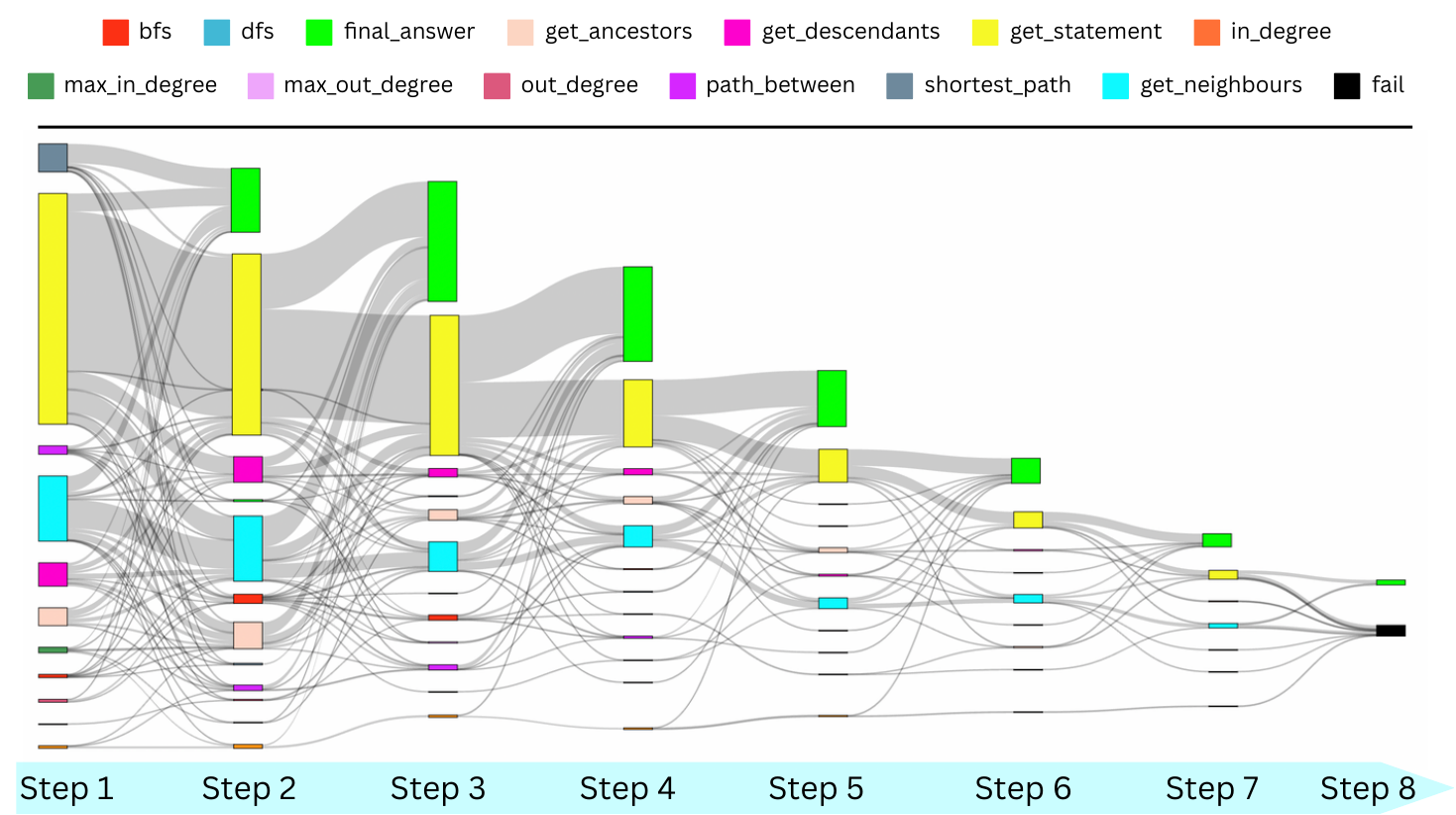}
    \caption{Flow diagram of the sequence of tools used by \texttt{FlowPathAgent} on \texttt{FlowExplainBench}. Each  \textit{Step} refers to a cycle of Tool Selection + Call.}
    \label{fig:sankey}
\end{figure}

Among the baselines, GPT4o Zero Shot Bounding Box shows the poorest performance. This model lacks inherent capabilities for mask generation, and instead generates bounding boxes in the textual token space, which is not well-suited for the task of flowchart attribution.

\begin{figure*}[ht]
    \centering
    \includegraphics[width=\linewidth]{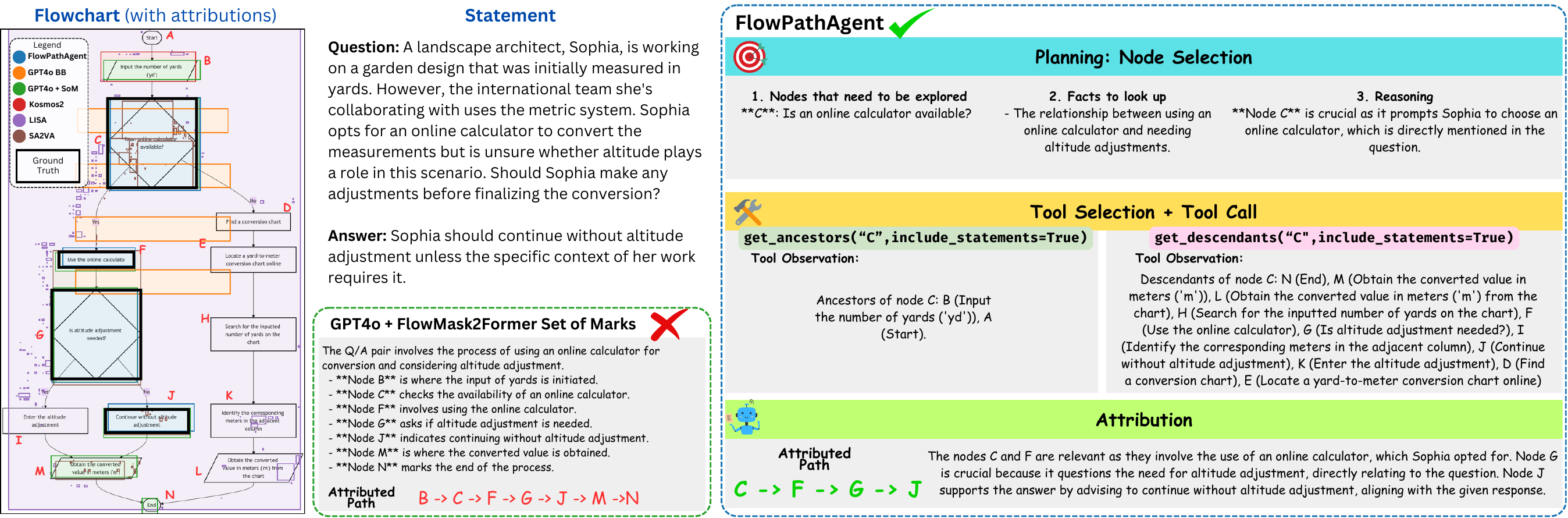}
    \caption{Qualitative comparison of \texttt{FlowPathAgent} with baseline methods. The flowchart illustrates attributions generated by various baselines, highlighting the agentic trace of \texttt{FlowPathAgent}. We contrast its output with the next strongest baseline, GPT-4o+SoM, to showcase differences in attribution quality and interpretability.}.
    \label{fig:qual}
\end{figure*}
In contrast, GPT4o SoM achieves a comparatively stronger performance. This can be attributed to the effective segmentation abilities of FlowMask2Former, which ensures that the elements to be attributed are accurately captured in the candidate set available to the model. Additionally, the reasoning capabilities of GPT4o contribute to improved performance by leveraging these segmented components in a logical manner. VisProg relies on weaker visual back-bones which do not understand images with text, leading to low detection rates, including none in FEBench-Wiki.

Further analysis of performance trends reveals an interesting behavior when we examine the performance of different models against the number of nodes in the flowchart, as illustrated in Fig \ref{fig:longtail}. As the complexity of the flowchart increases (i.e., as the number of nodes decreases), a performance dip is observed across all methods. This is likely due to the increasing difficulty in processing larger, more complex flowcharts, especially for models relying heavily on the visual presentation of the flowchart. In contrast, \texttt{FlowPathAgent} maintains a more consistent performance, with a relatively smaller dip in performance despite the increased complexity. This can be attributed to the model's ability to treat flowchart elements as logical entities, rather than solely relying on their visual representation. By leveraging its neurosymbolic approach, \texttt{FlowPathAgent} is able to more effectively process and attribute complex flowchart structures, providing robust and reliable attributions even in the long tail of node distributions.



\noindent\textbf{Qualitative Analysis.} Fig \ref{fig:qual} presents a qualitative comparison between \texttt{FlowPathAgent} and various baseline models. The GPT4o Zero Shot Bounding Box baseline fails to generate bounding boxes that overlap with or match the shape and dimensions of any flowchart nodes. On the other hand, LISA tends to overgeneralize by attributing the entire flowchart image, producing small, noisy masks that cover irrelevant areas, which reduces the clarity and precision of its attributions. Kosmos-2 also struggles with segmenting the nodes associated with the statement; it segments a single irrelevant node. SA2VA, while performing better than the other visual grounding models, still exhibits limitations. It generates low IoU masks around some correct nodes. Additionally, it sometimes produces extraneous masks that are not relevant to the flowchart’s logical structure. GPT4o with SoM shows some improvement, but tends to over-attribute by including steps that are further ahead in the flowchart than necessary. In contrast, \texttt{FlowPathAgent} excels by accurately detecting and attributing the entire flowchart path, identifying all the relevant nodes with high precision. Figure \ref{fig:sankey} displays the sequence of tools the agent employs across tool selection and execution steps, capped at 8 steps. The frequent use of the \texttt{get\_statement} tool highlights its vital role in verifying fact retrieval and scenario-based QA pairs without relying on visual input. Notably, Step 3 emerges as the most common final stage (evident by \texttt{final\_answer}), with nearly half of the penultimate tool calls dedicated to analyzing the graph structure. Additional agent behavior analysis and qualitative examples are provided in the Appendix.


\begin{table}[htbp]
\centering
\resizebox{\columnwidth}{!}{%
\begin{tabular}{lcccc}
\toprule
\multicolumn{1}{c}{\textbf{FlowMask2Former}} & \multicolumn{1}{c}{\textbf{FlowMask2Former}} & \multicolumn{1}{c}{\textbf{Flow2Mermaid VLM}} & \textbf{Attribution} & \textbf{\# of} \\
\multicolumn{1}{c}{\textbf{Performance Bucket}} & \multicolumn{1}{c}{\textbf{Avg. IoU (\%)}} & \multicolumn{1}{c}{\textbf{Performance (\%)}} & \textbf{F1 (\%)} & \textbf{Samples} \\
\midrule
(54.8, 63.2]  & \heatcell{59.71}{59.71}{91.56} & \heatcell{75.75}{75.75}{90.11} & \heatcell{56.25}{56.25}{86.65} & 8 \\
(63.2, 71.5]  & \heatcell{67.83}{59.71}{91.56} & \heatcell{82.54}{75.75}{90.11} & \heatcell{82.66}{56.25}{86.65} & 13 \\
(71.5, 79.8]  & \heatcell{76.98}{59.71}{91.56} & \heatcell{86.13}{75.75}{90.11} & \heatcell{83.64}{56.25}{86.65} & 63 \\
(79.8, 88.1]  & \heatcell{84.98}{59.71}{91.56} & \heatcell{88.86}{75.75}{90.11} & \heatcell{86.65}{56.25}{86.65} & 364 \\
(88.1, 96.4]  & \heatcell{91.56}{59.71}{91.56} & \heatcell{90.11}{75.75}{90.11} & \heatcell{84.27}{56.25}{86.65} & 477 \\
\bottomrule
\end{tabular}
}
\caption{Binned analysis shows that while segmentation quality (IoU) marginally influences Word Overlap F1, the overall task F1 remains relatively stable—indicating limited error propagation across pipeline stages.}
\label{tab:err}
\end{table}

\begin{figure}
    \centering
    \includegraphics[width=0.7\linewidth]{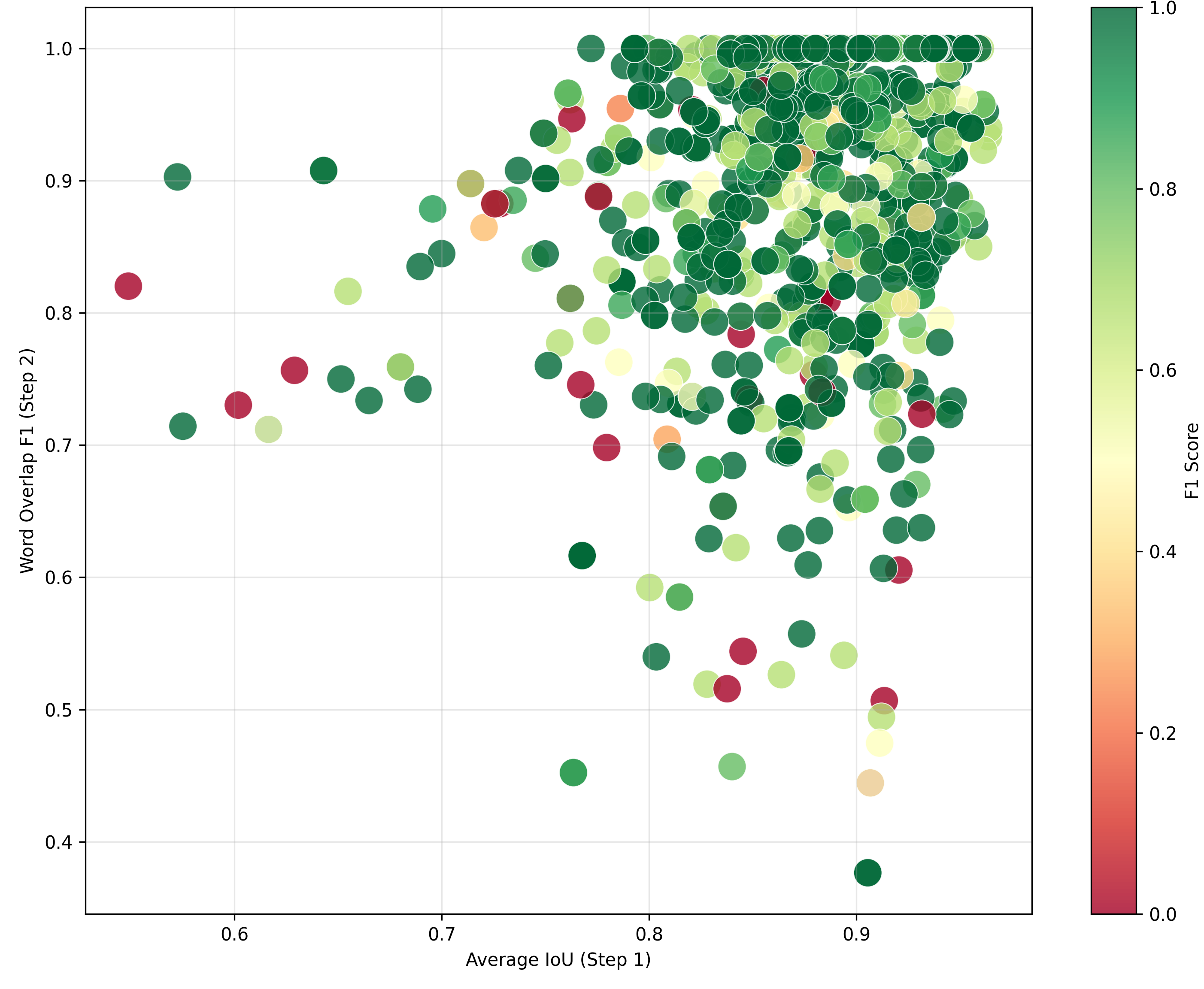}
    \caption{Scatter plot of segmentation IoU versus Word Overlap F1 for individual samples, color-coded by overall task F1. Clustering in the high-performance region indicates minimal error propagation across the pipeline.}

    \label{fig:err}
\end{figure}

\noindent\textbf{Error Propagation}: An inherent limitation in modular agentic systems is that inaccuracies in one component can affect downstream results. On the full benchmark dataset, FlowMask2Former achieved a high overall Jaccard Similarity (IoU > 0.5) of 0.98, and Flow2Mermaid VLM obtained a Word F1 score of 0.89. To better understand the relationship between segmentation quality and transcription fidelity, we performed a binned analysis in Table~\ref{tab:err}, grouping samples by segmentation IoU. Complementing this, Fig.~\ref{fig:err} visualizes the correlation between IoU and Word Overlap F1 for individual data points, colored by overall task F1. The concentration of points with high IoU and high Word Overlap F1 alongside consistently strong task performance suggests limited error propagation across the pipeline. This stability likely arises from the high accuracy of FlowMask2Former and Flow2Mermaid VLM, as well as the neurosymbolic agent’s role in verifying and correcting errors, enabling the modules to complement each other and reduce cascading failures.

\section{Conclusion}
We introduced the task of Flowchart Attribution and proposed \texttt{FlowExplainBench} for evaluating fine-grained visual grounding in flowchart QA. We presented \texttt{FlowPathAgent}, a neurosymbolic agent that leverages graph-based reasoning to accurately identify the optimal path underpinning LLM responses. Experimental results demonstrate significant improvements over existing baselines, highlighting the effectiveness of our approach. 

\section{Limitations}

While our approach demonstrates strong performance, there are areas for further improvement. First, although \texttt{FlowPathAgent} effectively integrates symbolic reasoning, it builds on FlowMask2Former for segmentation and Flow2Mermaid VLM for converting visual flowcharts to mermaid code. As with any modular system, potential errors in these components may influence overall performance. However, our framework remains flexible, allowing for seamless integration of alternative models better suited to specific scenarios.

Second, our benchmark, \texttt{FlowExplainBench}, captures a diverse range of flowchart structures but does not yet encompass all real-world variations, such as hand-drawn diagrams. The primary challenge lies in the availability of high-quality datasets with comprehensive annotations. While existing methods address hand-drawn flowchart segmentation, scaling them for attribution remains an open area of research. Future work could explore semi-supervised or automated annotation strategies to enhance coverage.

Lastly, our approach is designed for static flowcharts, and extending it to dynamic or interactive systems presents an opportunity for further research. Many real-world applications involve evolving decision-making processes, which could benefit from models that handle sequential updates and conditional dependencies. 

Future work could address these limitations by improving segmentation robustness, expanding the benchmark to include more diverse flowchart types, and developing models capable of handling dynamic and interactive flowcharts. Additionally, integrating reinforcement learning or self-supervised learning techniques could enhance model adaptability and generalization across various flowchart formats.

\section{Ethics Statement}
In this study, we utilize the publicly accessible FlowVQA dataset, which is distributed under the MIT License\footnote{\url{https://github.com/flowvqa/flowvqa?tab=MIT-1-ov-file}}. We ensure that the identities of human evaluators remain confidential, and no personally identifiable information (PII) is used at any stage of our research. This work is focused exclusively on applications for fine-grained visual flowchart attribution and is not intended for other use cases. We also recognize the broader challenges associated with large language models (LLMs), including potential risks related to misuse and safety, and we encourage readers to consult the relevant literature for a more detailed discussion of these issues \citep{risks1, risks2, risks3}.

\bibliography{custom}

\appendix

\section{Further Details}

\subsection{Baselines}
\label{baselines}
\textbf{Zero-shot GPT-4o Bounding Box} We use GPT-4o \cite{openai_gpt4o_2024} to predict normalized bounding box coordinates for chart components based on text and the visual chart, following established methods for zero-shot localization \cite{yang2023dawn}.

\noindent\textbf{Kosmos-2:}\cite{peng2023kosmos} is a multimodal large language model that combines text-to-visual grounding, supporting tasks like referring expression interpretation and bounding box generation by linking objects in images with text.

\noindent\textbf{LISA:} \cite{li2023llatrieval} is a model for generating segmentation masks from textual queries, extending VLM capabilities to segmentation tasks, and excels in zero-shot performance with minimal fine-tuning on task-specific data.

\noindent\textbf{SA2VA}\cite{yuan2025sa2vamarryingsam2llava} is a unified model for dense grounded understanding of both images and videos, combining SAM-2 for segmentation and LLaVA for vision-language tasks, enabling robust performance in referring segmentation.

\noindent\textbf{VisProg} \cite{Gupta2022VisProg} is an agent that generates interpretable visual programs by decomposing queries into executable steps, enabling modular and explainable visual reasoning.

\noindent\textbf{GPT4o + FlowMask2Former SoM Prompting,} as an ablation study, we incorporate this baseline where GPT-4o utilizes the segmented flowchart generated by FlowMask2Former, and applies Set-of-Marks (SoM) \cite{yang2023set} prompting to guide the model's predictions.

\subsection{Implementation Details}
\label{implementation_details}

The \texttt{facebook/mask2former-swin-tiny-coco} \texttt{-instance} model is fine-tuned for 20 epochs for FlowMask2Former, employing a learning rate of \(1 \times 10^{-5}\) with a cosine annealing scheduler. A batch size of 4 is used, and gradient accumulation occurs over 4 steps to address memory constraints. Training is conducted using 16-bit precision to improve computational efficiency. In the case of the Mermaid2Graph Vision-Language Model (VLM), fine-tuning is performed on the \texttt{unsloth/Qwen2-VL-7B-Instruct} checkpoint, focusing on vision, language, attention, and MLP layers. This model is trained for 3 epochs with a batch size of 1 and gradient accumulation over 5 steps. A learning rate of \(2 \times 10^{-4}\) is applied with a linear scheduler. To optimize memory usage, the model is loaded in 4-bit precision, and AdamW is used as the optimizer with a weight decay of 0.01. The baseline models are initialized as follows: LISA from \texttt{xinlai/LISA-13B-llama2-v1}, Kosmos-2 from \texttt{microsoft/kosmos-2-patch14-22}, and Sa2VA from \texttt{ByteDance/Sa2VA-8B}. Default settings and parameters are used for all baselines.

\subsection{Computational Budget}

Table \ref{tab:comp_budget_with_gpu} shows the computational budget for this paper, broken down by associated tasks.
\begin{table}[ht]
\centering
\resizebox{\columnwidth}{!}{%
\begin{tabular}{|l|l|l|l|}
\hline
\textbf{Task} & \textbf{Time (hours)} & \textbf{\# of GPUs} & \textbf{GPU Spec} \\ \hline
FlowMask2Former Training & 14 & 1 & NVIDIA RTX A6000 \\ \hline
Flow2Mermaid VLM Training & 8 & 1 & NVIDIA RTX A6000 \\ \hline
Baseline, Trained Model Inference & 3 & 1 & NVIDIA RTX A6000 \\ \hline
\end{tabular}%
}
\caption{Computational Budget for experiments in the paper}
\label{tab:comp_budget_with_gpu}
\end{table}

\begin{figure}[ht]
    \centering
    \includegraphics[width=\linewidth]{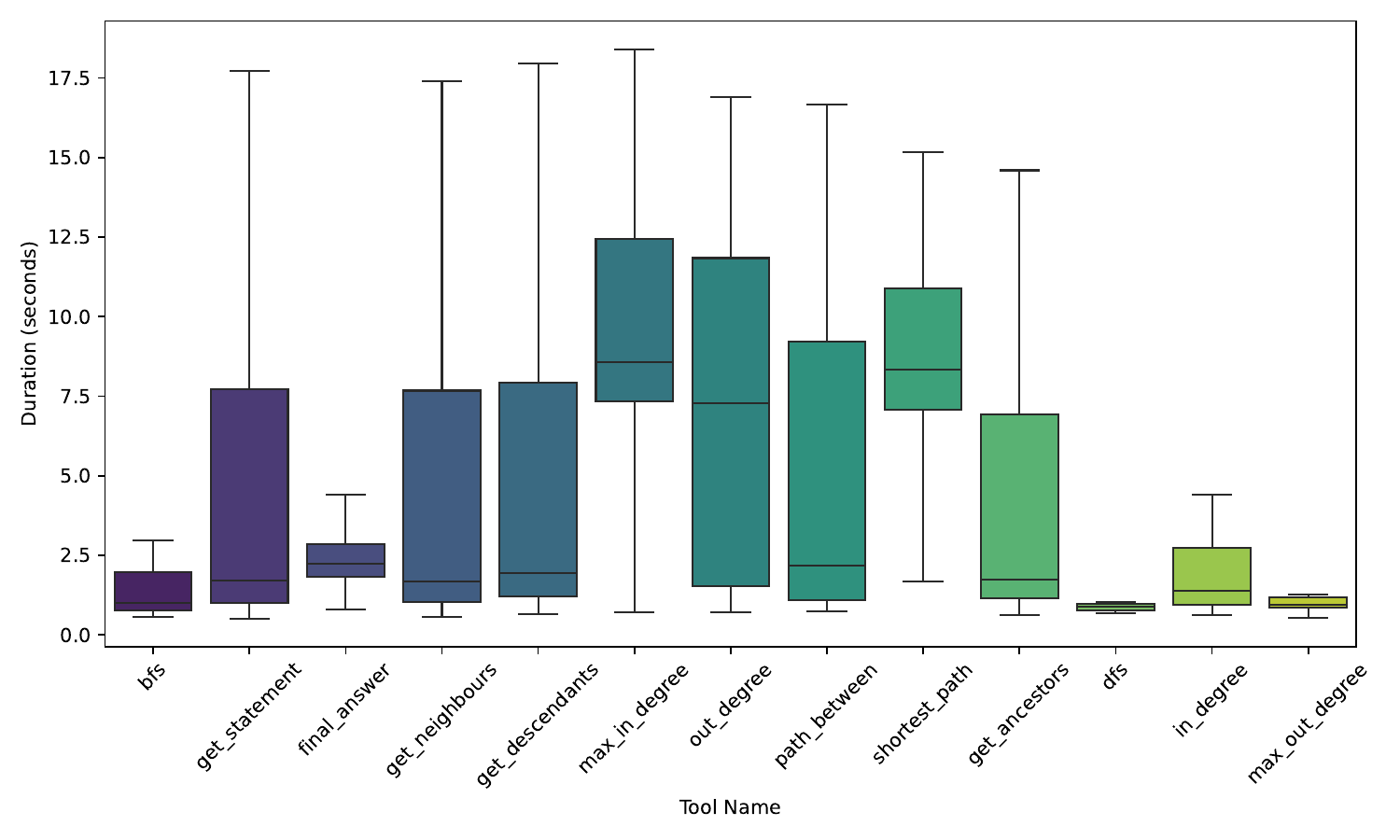}
    \caption{Box-plot distribution of time taken in each tool call, in seconds.}
    \label{fig:tool_time}
\end{figure}

\begin{figure}
    \centering
    \includegraphics[width=\linewidth]{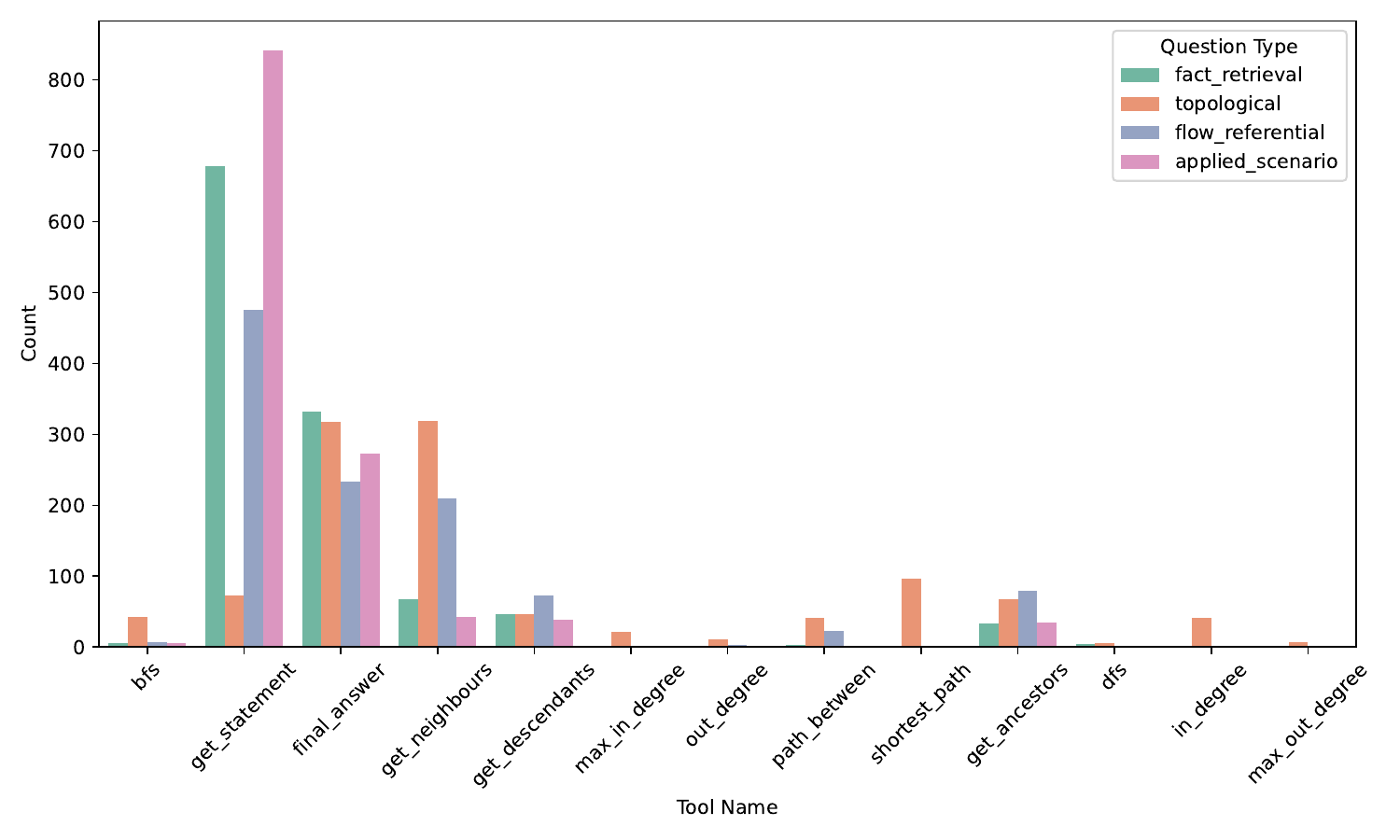}
    \caption{Distribution of count of tool calls, segregated by question type.}
    \label{fig:tool_question}
\end{figure}
\subsection{Dataset}
\begin{figure}
    \centering
    \includegraphics[width=\linewidth]{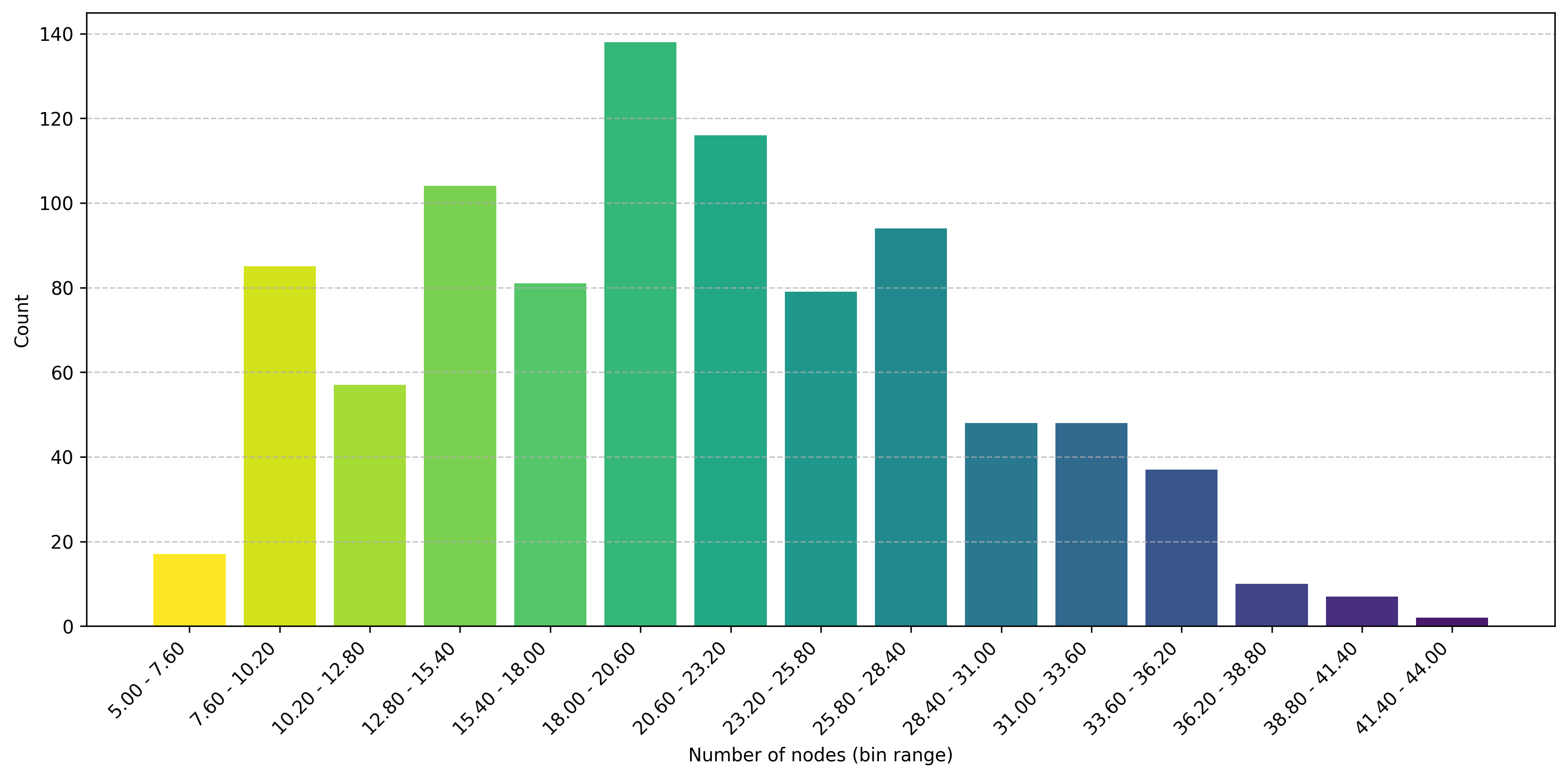}
    \caption{Distribution of nodes in our benchmark.}
    \label{fig:node_dist}
\end{figure}

Fig~\ref{fig:node_dist} shows the distribution of nodes in our benchmark.

\begin{figure*}[ht]
    \centering
    \includegraphics[width=\linewidth]{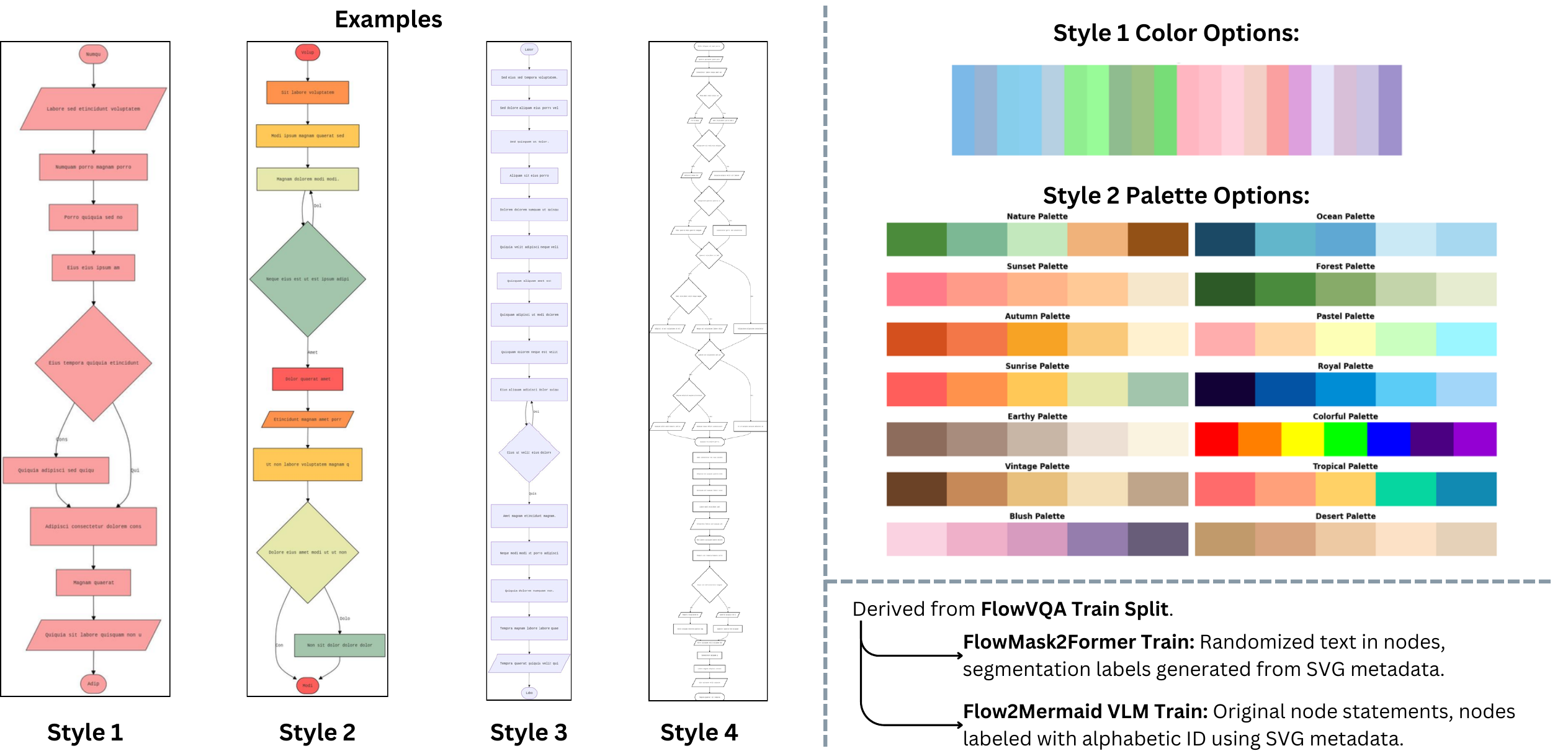}
    \caption{Overview of training split used for FlowMask2Former, and Flow2Mermaid VLM. The figure demonstrates the style options, color palettes used, and distinction between both training sets.}
    \label{fig:example_train}
\end{figure*}

Fig \ref{fig:example_train} represents examples from the training set for FlowMask2Former and Flow2Mermaid VLM, displaying different style types.

\begin{figure}
    \centering
    \includegraphics[width=\linewidth]{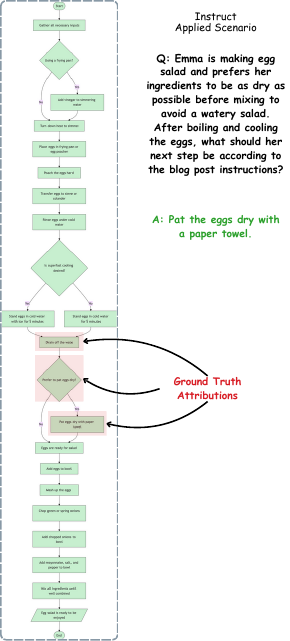}
    \caption{Example from \texttt{FlowExplainBench--Instruct}. This example represents an \textit{Applied Scenario} question, and has a style type 1 (single color).}
    \label{fig:gt_eg1}
\end{figure}

\begin{figure}
    \centering
    \includegraphics[width=\linewidth]{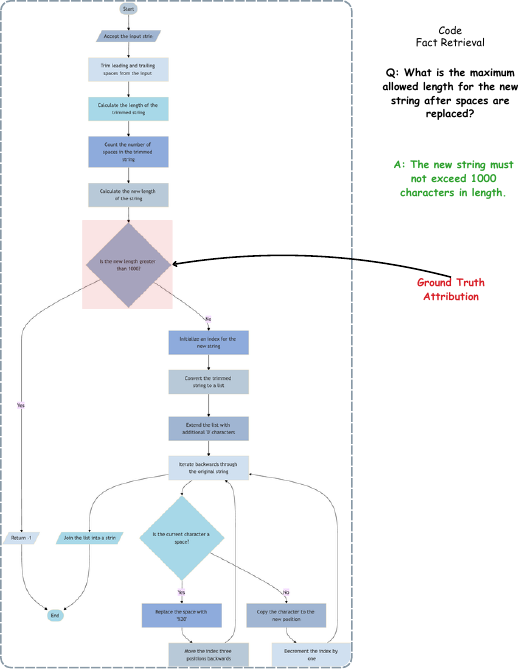}
    \caption{Example from \texttt{FlowExplainBench--Code}. This example represents a \textit{Fact Retrieval} question, and has a style type 2 (multiple colors).}
    \label{fig:gt_eg2}
\end{figure}

\begin{figure}
    \centering
    \includegraphics[width=\linewidth]{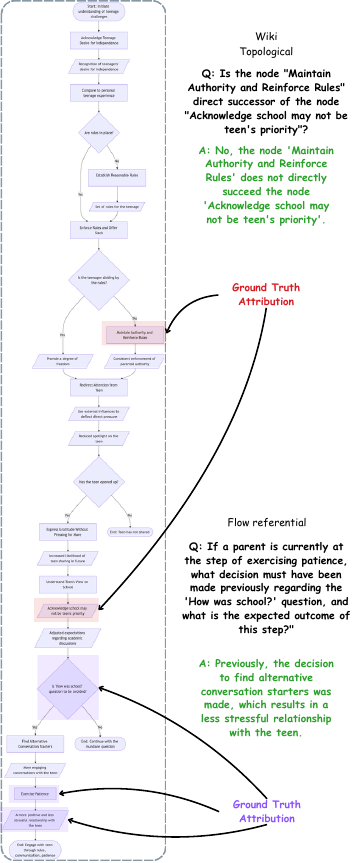}
    \caption{Example from \texttt{FlowExplainBench--Instruct}. This example represents  \textit{Tolopolgical} and \textit{Flow Referential} questions, and has a style type 3 (mermaid default).}
    \label{fig:gt_eg3}
\end{figure}

Figs \ref{fig:gt_eg1}-\ref{fig:gt_eg3} represent examples from \texttt{FlowExplainBench} from different domains, with different styles and question types.

\begin{figure}
    \centering
    \includegraphics[width=\linewidth]{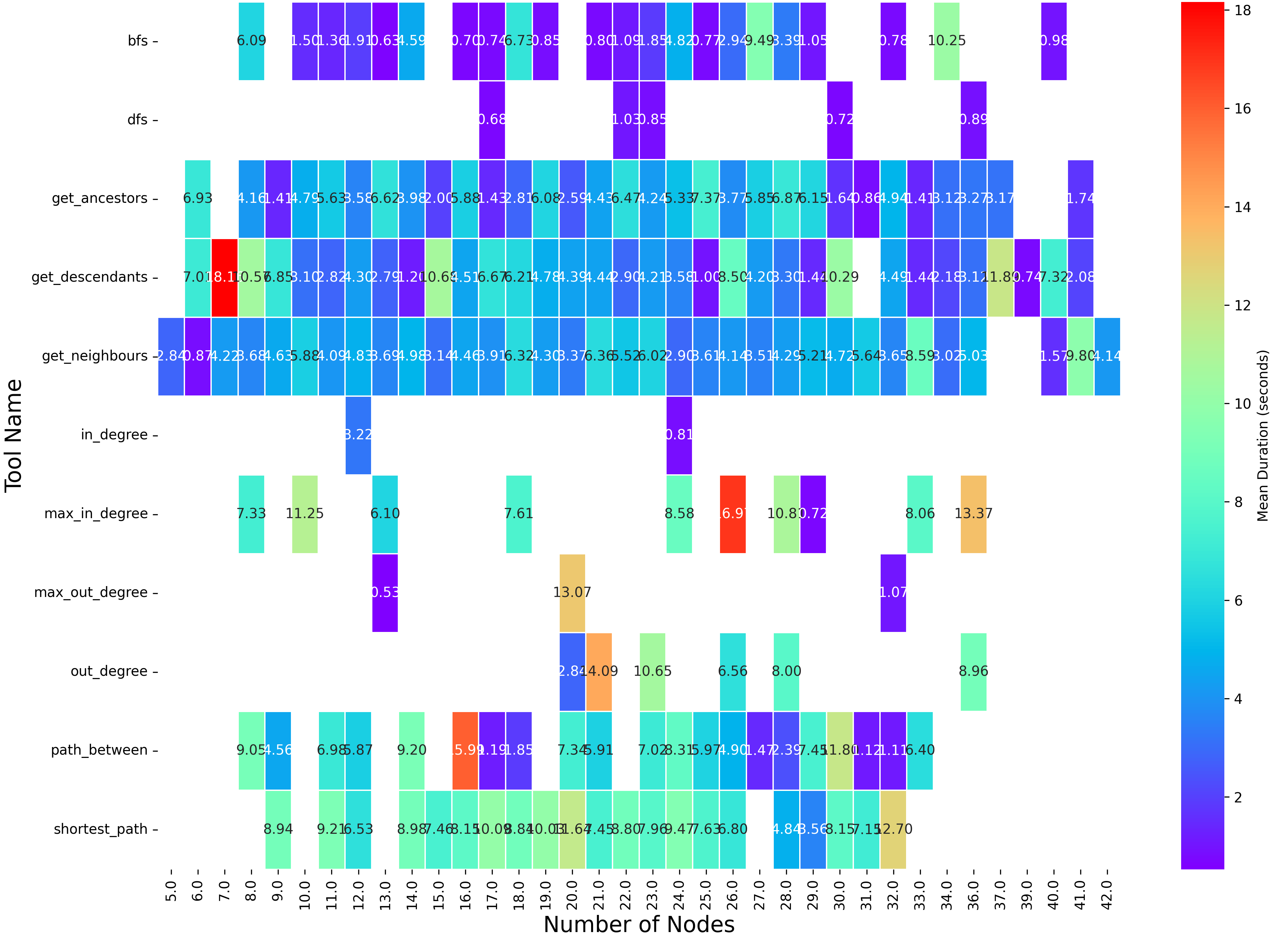}
    \caption{Heatmap of time taken by different tools to execute, binned by number of nodes in the flowchart.}
    \label{fig:tool_time_heat}
\end{figure}

\begin{figure}
    \centering
    \includegraphics[width=\linewidth]{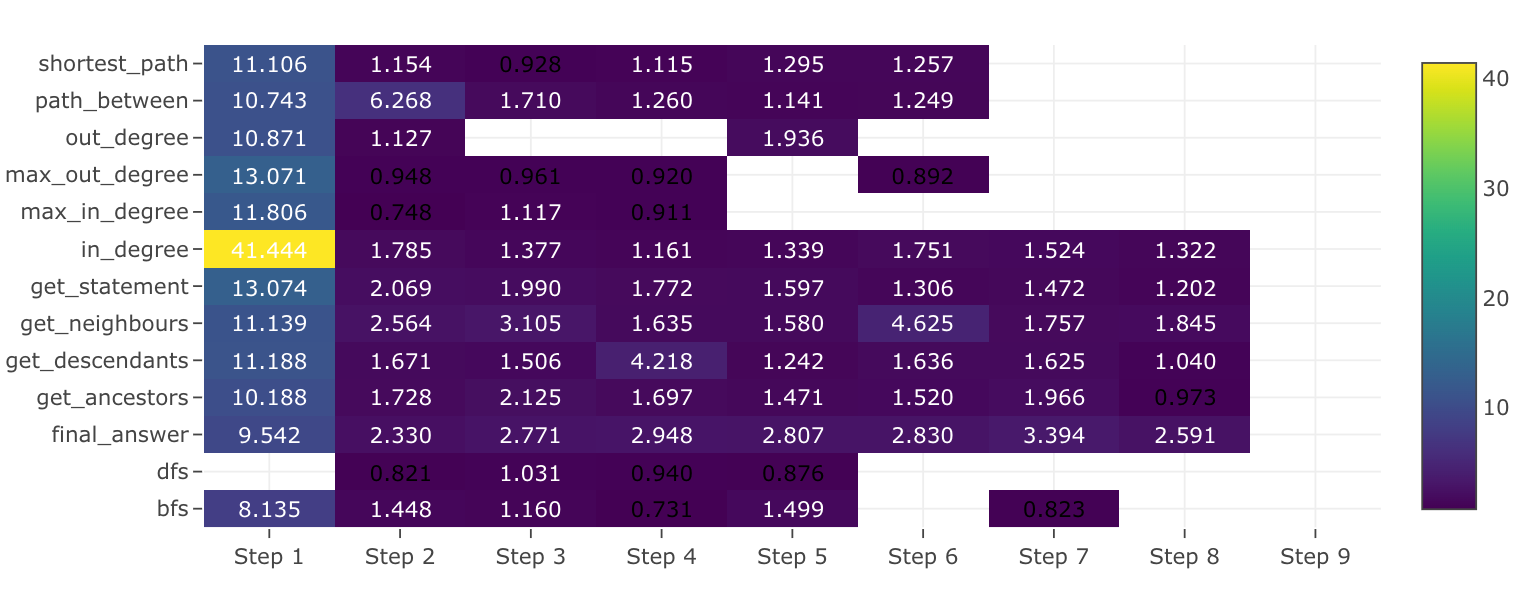}
    \caption{Heatmap of duration of tool call execution, arranged by agentic step.}
    \label{fig:step_time_heat}
\end{figure}
\begin{table*}[]
    \centering
    \resizebox{1.5\columnwidth}{!}{%
    \begin{tabular}{lll}
    \toprule
    Name & Worst-Case Time Complexity & Justification \\
    \midrule
    get\_statement & \(O(1)\) & Constant time operation regardless of graph size \\
    get\_ancestors & \(O(n^2)\) & In a complete graph, could require traversing all edges from all nodes \\
    get\_descendants & \(O(n^2)\) & Similarly, might need to explore all possible paths in a dense graph \\
    get\_neighbours & \(O(n)\) & In worst case, a node could be connected to all other nodes \\
    in\_degree & \(O(n)\) & May need to check all nodes to count incoming edges \\
    out\_degree & \(O(n)\) & May need to check all nodes to count outgoing edges \\
    max\_in\_degree & \(O(n)\) & Must examine every node to find maximum \\
    max\_out\_degree & \(O(n)\) & Must examine every node to find maximum \\
    bfs & \(O(n^2)\) & In a complete graph, each node has \(n-1\) edges, so total is \(O(n^2)\) \\
    dfs & \(O(n^2)\) & Same as BFS in worst case with dense graph \\
    path\_between & \(O(n^2)\) & May need to explore all possible paths in worst case \\
    shortest\_path & \(O(n^2)\) & BFS-based shortest path in a dense graph \\
    final\_answer & \(O(1)\) & Constant time operation \\
    \bottomrule
    \end{tabular}%
    }
    \caption{Time Complexity of tools used in our system.}
    \label{tab:time_complex}
\end{table*}

\subsection{Qualitative Examples}
Fig. \ref{fig:qual_2} and \ref{fig:qual_2} present qualitative comparisons of the baseline methods. While these examples do not comprehensively represent the overall performance ranking, they have been deliberately chosen to highlight specific limitations and failure cases of each method. This selection aims to provide insights into the scenarios where certain approaches struggle, offering a clearer understanding of their weaknesses.

\begin{figure*}
    \centering
    \includegraphics[width=\linewidth]{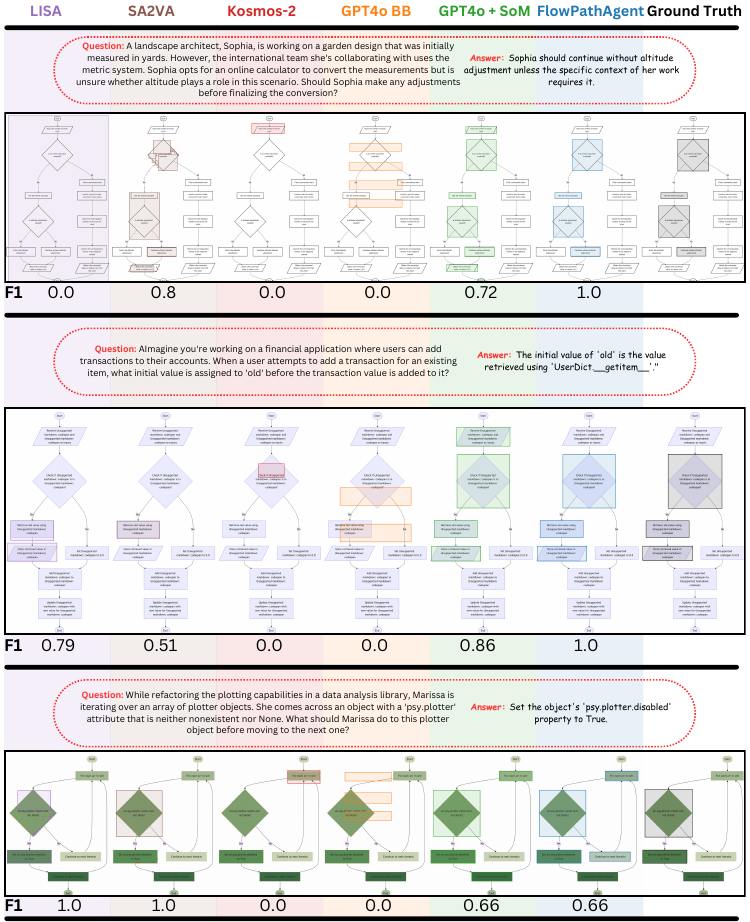}
    \caption{Qualitative comparison of \texttt{FlowPathAgent} with baselines via examples.}
    \label{fig:qual_3}
\end{figure*}

\begin{figure*}
    \centering
    \includegraphics[width=\linewidth]{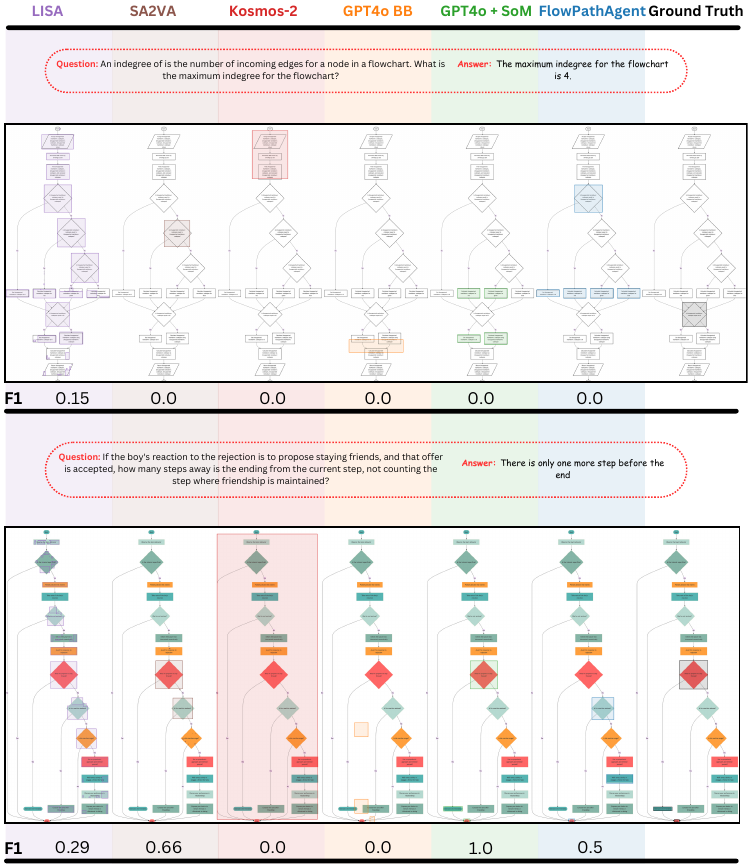}
    \caption{(Continued) Qualitative comparison of \texttt{FlowPathAgent} with baselines via examples.}
    \label{fig:qual_2}
\end{figure*}
\section{Agent Analysis}

\subsection{API Description}
\begin{figure*}
    \centering
    \includegraphics[width=0.5\linewidth]{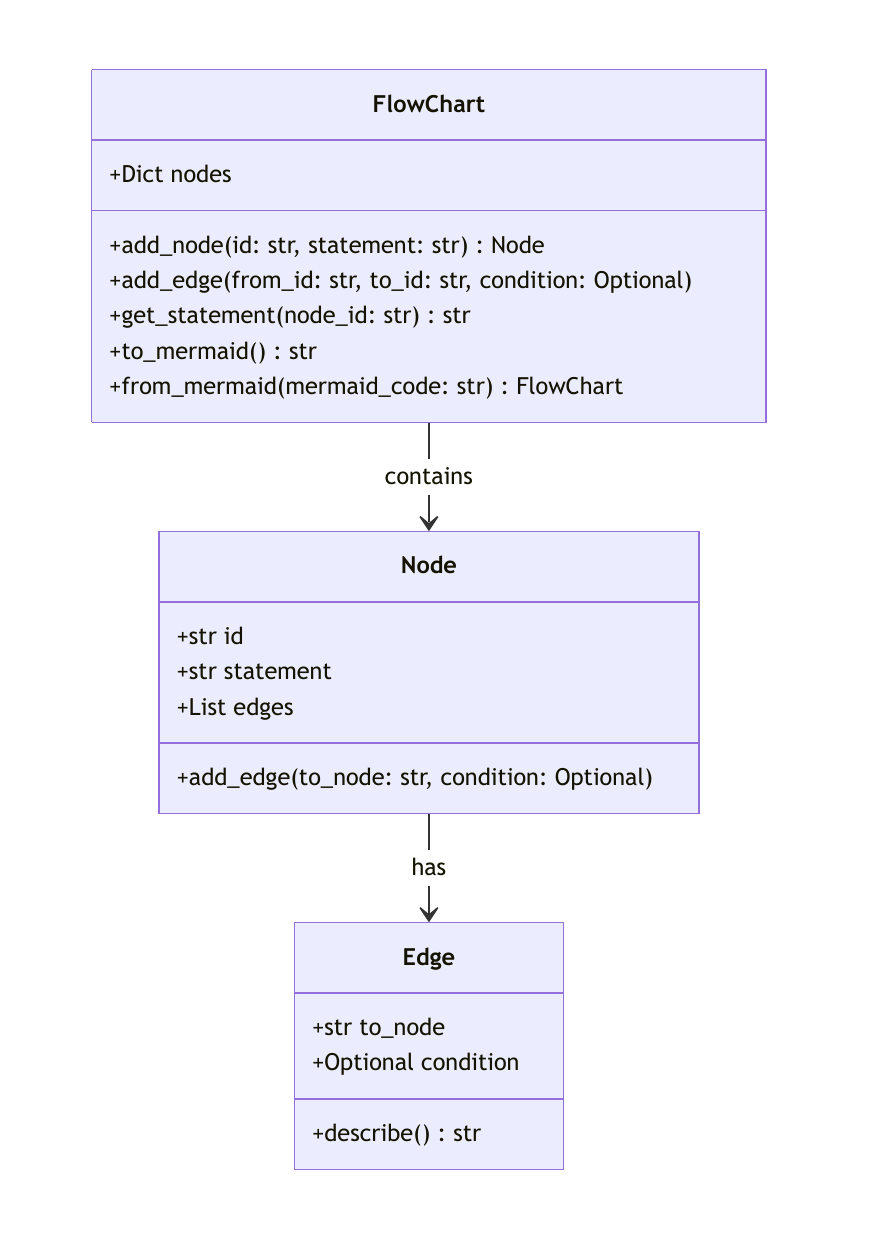}
    \caption{Class Diagram of the \texttt{FlowChart} data structure representing directed graphs with conditional edges.}
    \label{fig:API}
\end{figure*}

Fig. \ref{fig:API} shows the class diagram of the data structure used to represent Nodes, Edges, and the Flowchart. The \texttt{FlowChart} class serves as the primary structure, managing a collection of \texttt{Node} objects, each identified by a unique ID and containing a statement. Nodes are interconnected through \texttt{Edge} objects, which define directed relationships with optional conditions (\texttt{Yes}, \texttt{No}, or unconditional). 

Table \ref{tab:tools} summarises the API for the tools provided to \texttt{FlowPathAgent}. Except for \texttt{final\_answer} which returns the final answer and reasoning involved, all other tools operate on a global \texttt{FlowChart} object initiated from mermaid code generated by Flow2Mermaid VLM.

\begin{table*}[ht]
    \centering
    \small
    \begin{tabularx}{\textwidth}{|l|X|X|}
        \hline
        \textbf{Name} & \textbf{Description} & \textbf{Arguments} \\
        \hline
        get\_statement & Returns the statement associated with a node. & 
        node\_id (string): Identifier of the node. \\
        \hline
        get\_ancestors & Identifies all nodes that have paths leading to the specified node. & 
        node\_id (string): Identifier of the target node. \newline
        levels (integer, optional): Maximum levels to traverse. \newline
        include\_statements (boolean, optional): If True, includes statements. Defaults to False. \\
        \hline
        get\_descendants & Identifies all nodes that can be reached from the specified node. & 
        node\_id (string): Identifier of the starting node. \newline
        levels (integer, optional): Maximum levels to traverse. \newline
        include\_statements (boolean, optional): If True, includes statements. Defaults to False. \\
        \hline
        get\_neighbours & Returns all nodes connected to the given node by outgoing edges. & 
        node\_id (string): Identifier of the node. \newline
        include\_statements (boolean, optional): If True, includes statements. Defaults to False. \\
        \hline
        in\_degree & Returns the number of incoming edges to a node. & 
        node\_id (string): Identifier of the node. \\
        \hline
        out\_degree & Returns the number of outgoing edges from a node. & 
        node\_id (string): Identifier of the node. \\
        \hline
        max\_in\_degree & Identifies nodes with the highest incoming edges. & None \\
        \hline
        max\_out\_degree & Identifies nodes with the highest outgoing edges. & None \\
        \hline
        bfs & Performs breadth-first search from a starting node. & 
        start\_id (string, optional): Identifier of the node. \newline
        conditions (object, optional): Dictionary of edge conditions. \newline
        include\_statements (boolean, optional): If True, includes statements. Defaults to False. \\
        \hline
        dfs & Performs depth-first search from a starting node. & 
        start\_id (string, optional): Identifier of the node. \newline
        conditions (object, optional): Dictionary of edge conditions. \newline
        include\_statements (boolean, optional): If True, includes statements. Defaults to False. \\
        \hline
        path\_between & Finds a path between two nodes, considering edge conditions. & 
        start\_id (string): Start node. \newline
        end\_id (string): End node. \newline
        conditions (object, optional): Edge conditions. \newline
        include\_statements (boolean, optional): If True, includes statements. Defaults to False. \\
        \hline
        shortest\_path & Finds the shortest path between two nodes using BFS. & 
        start\_id (string): Start node. \newline
        end\_id (string): End node. \newline
        conditions (object, optional): Edge conditions. \newline
        include\_statements (boolean, optional): If True, includes statements. Defaults to False. \\
        \hline
        final\_answer & Provides a final answer to the given problem. & 
        answer (any): The final answer. \\
        \hline
    \end{tabularx}
    \caption{Tools provided to \texttt{FlowPathAgent}.}
    \label{tab:tools}
\end{table*}

\subsection{Tool-use Analysis}

Fig \ref{fig:tool_time} shows the distribution of run-time of tool cals, called from within the agentic framework. \texttt{max\_in\_degree} has the maximum median run-time, which can be explained by the fact that the \texttt{Node} data-structure employed by us only has outgoing edges, meaning all nodes have to be iterated to find the node with maximum in-degree. The second highest median run-time belongs to \texttt{shortest\_path}, which is implemented as a O(V+E) breadth-first-search based algorithm. Table~\ref{tab:time_complex} describes the theoretical time complexity of the tool calls. Fig~\ref{fig:tool_time_heat} represents the time of execution by tool, as a heatmap plotted along the number of nodes. As seen from this heatmap,  in practice, the constant compute associated with each tool call often outweighs the cost incurred by increasing the number of nodes, when the number of nodes is not large enough. Moreover, fig~\ref{fig:step_time_heat}, which plots the time taken by each tool to execute by agentic step shows that highest latencies occur for the first step of the simulation, because of compute cost of initialization.

We analyze the distribution of tool calls by question type in Fig \ref{fig:tool_question}. Topological questions show the most diversity in terms of tool calls, since they require interpreting structural aspects of the FlowChart.  \texttt{get\_statement} is the most common tool for the other question types. This is because all the other questions require content from inside the flowchart, and often involve multiple calls of \texttt{get\_statement} in a single agentic run. For flow-referential questions, \texttt{get\_neighbours} is a popular tool, since this tool allows downstream flow analysis from an anchor node.

\subsection{Prompts and Implementation}
\begin{figure}
    \centering
    \includegraphics[width=\linewidth]{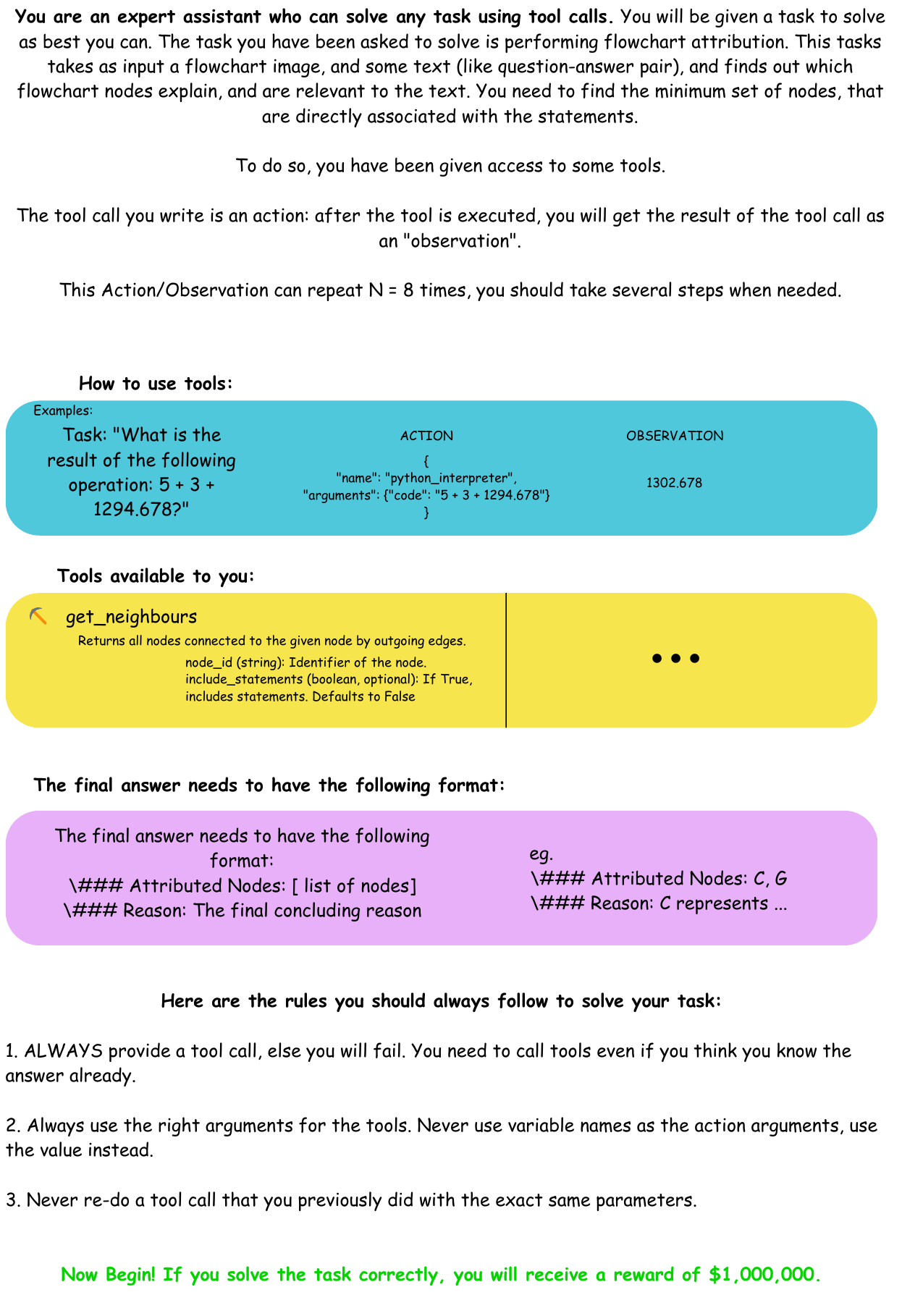}
    \caption{System prompt template provided to \texttt{FlowPathAgent}.}
    \label{fig:system_prompt}
\end{figure}

\begin{figure}
    \centering
    \includegraphics[width=\linewidth]{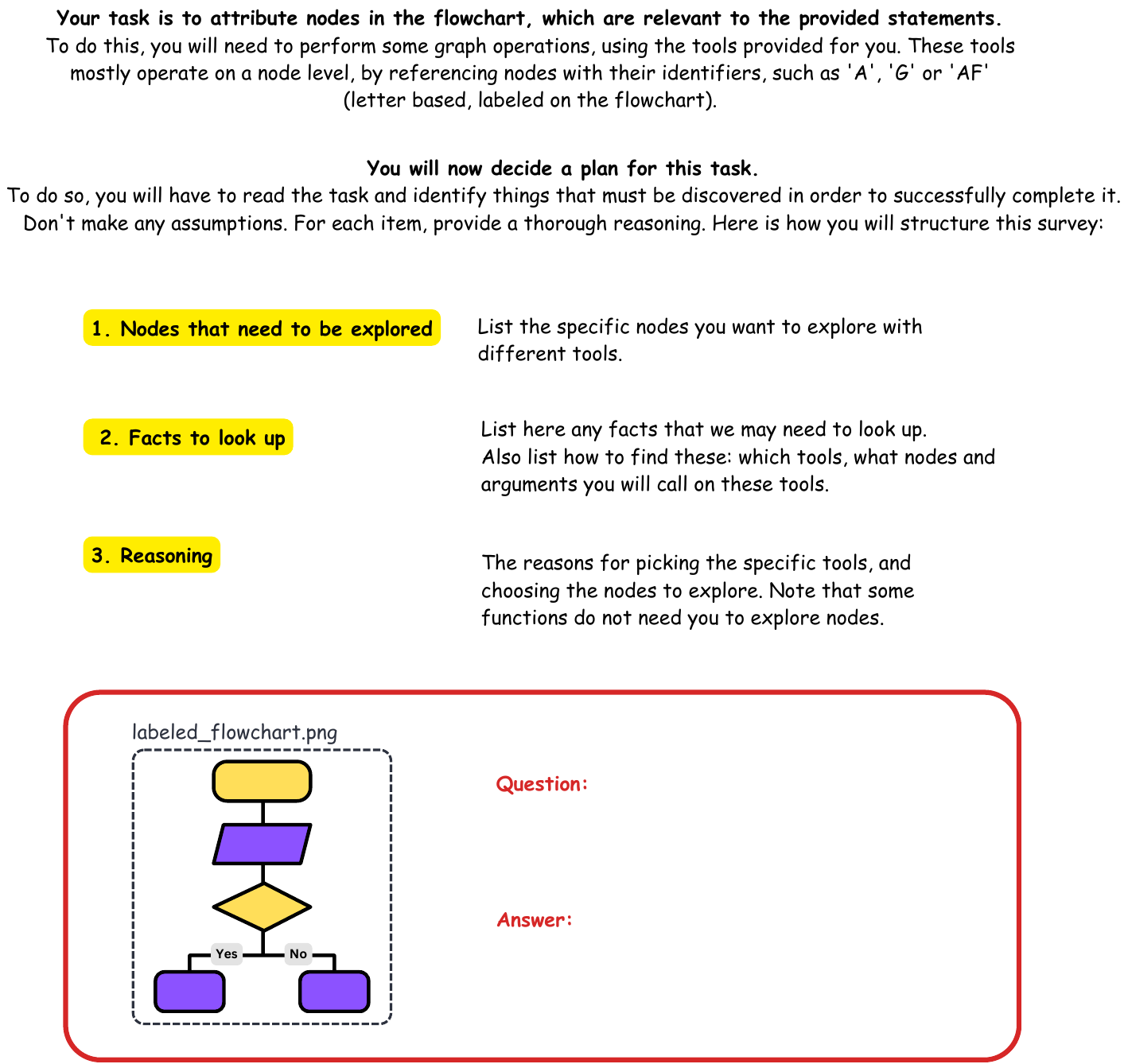}
    \caption{Planning prompt template provided to \texttt{FlowPathAgent}.}
    \label{fig:planning_prompt}
\end{figure}

Fig \ref{fig:system_prompt} and \ref{fig:planning_prompt} represent the system prompt template, and planning prompt template used by \texttt{FlowPathAgent}. We implemented \texttt{FlowPathAgent} using HuggingFace's smolagents \footnote{\url{https://github.com/huggingface/smolagents}} library. We patched the library to ensure that visual tokens are only used in the planning step (node selection step), and removed from the conversation template thereafter.

\section{Benchmark Construction}
\subsection{Automatic Labeling}
\begin{figure}
    \centering
    \includegraphics[width=\linewidth]{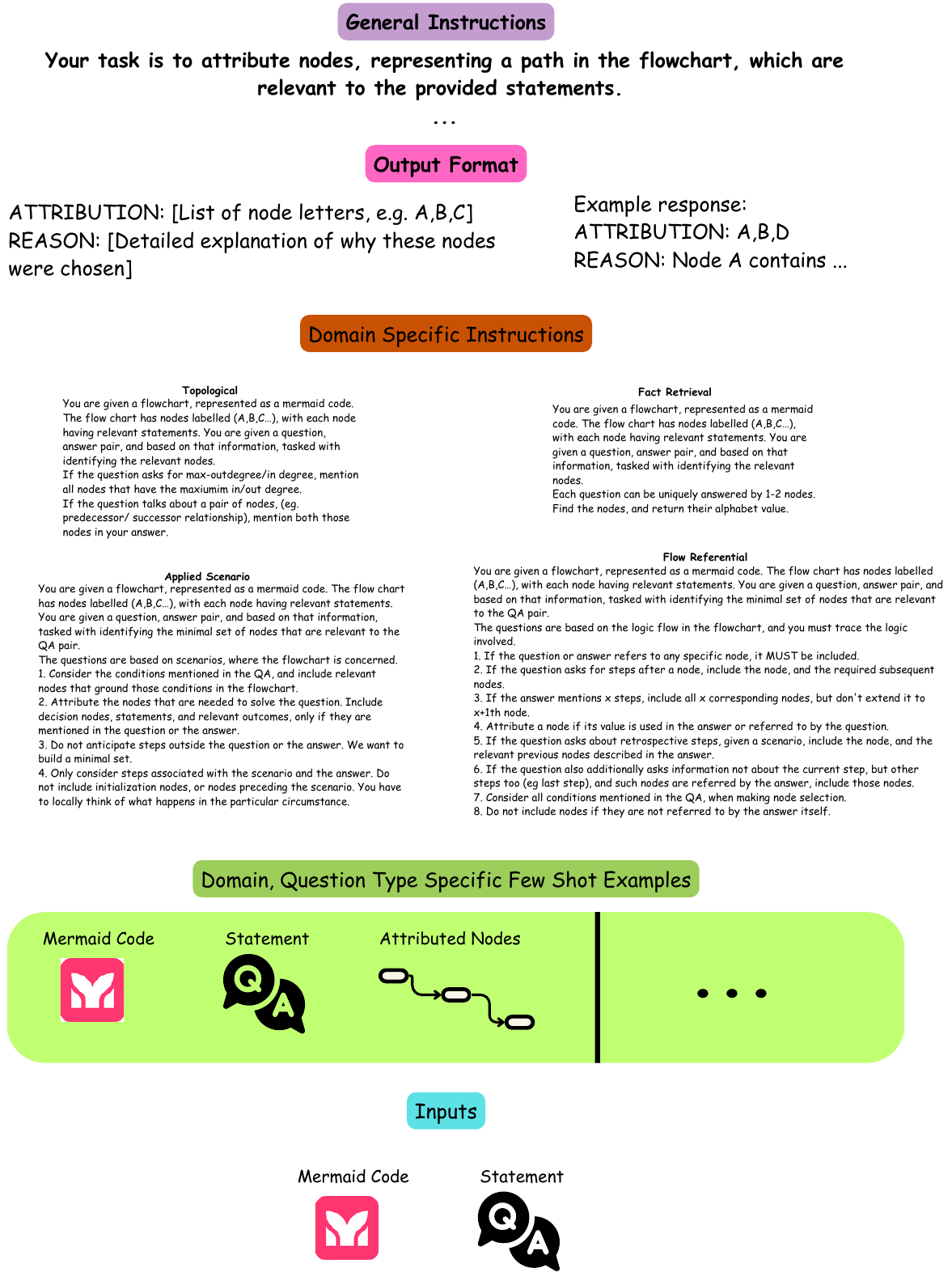}
    \caption{Prompt Template used for initial automatic ground truth annotation using GPT4.}
    \label{fig:automatic_prompt}
\end{figure}

Fig \ref{fig:automatic_prompt} represents the prompt template used to perform automatic annotations, in step 1 of our ground truth annotation process.

\subsection{Style Diversity}
\begin{figure}
    \centering
    \includegraphics[width=\linewidth]{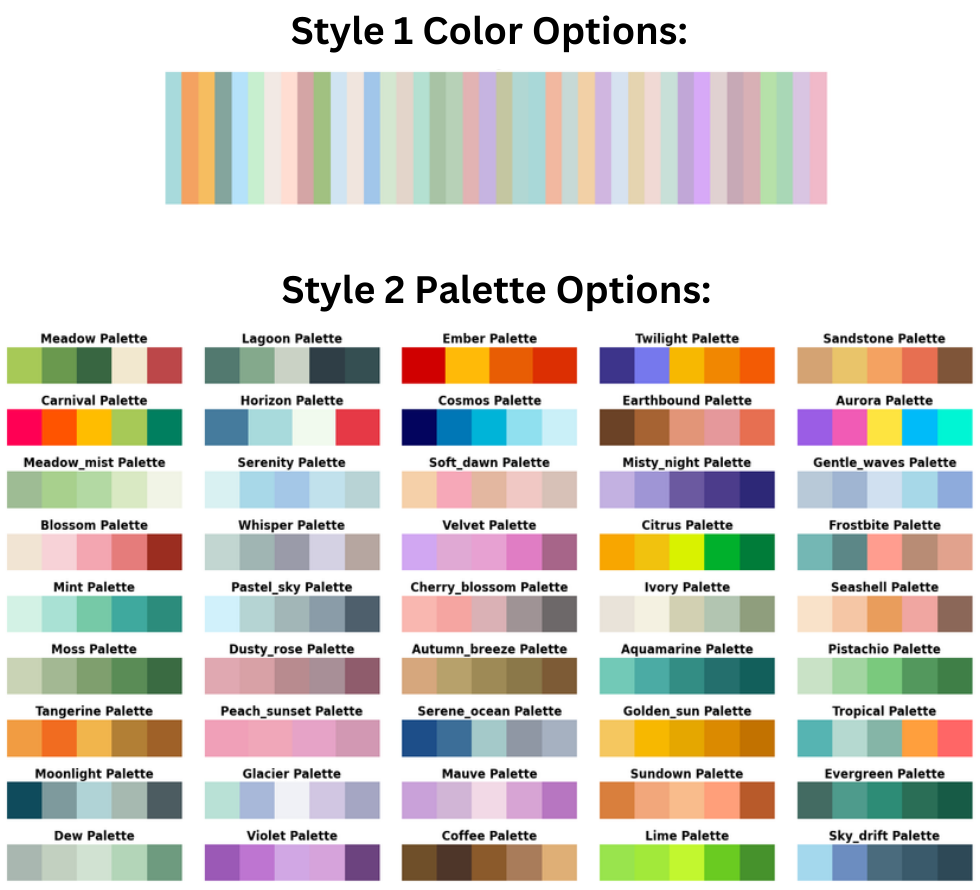}
    \caption{Diversity of color schemes used to augment \texttt{FlowExplainBench} flowchart styles}
    \label{fig:test_color_schemes}
\end{figure}

Fig \ref{fig:test_color_schemes} represents the color schemes used to augment style diversity in \texttt{FlowExplainBench}. Styles for auxiliary datasets used in this paper are presented in Fig \ref{fig:example_train}. 
\subsection{Human Annotation}
\label{annotation}
We employed two graduate student annotators, aged 22-25. The annotators were proficient in English, and were exposed to flowchart QA samples from the training set before the annotation exercise, to make them comfortable with the flowcharts involved.  The annotators were fairly compensated at the standard Graduate Assistant hourly rate, following their respective graduate school policies.  

\begin{figure*}
    \centering
    \includegraphics[width=0.8\linewidth]{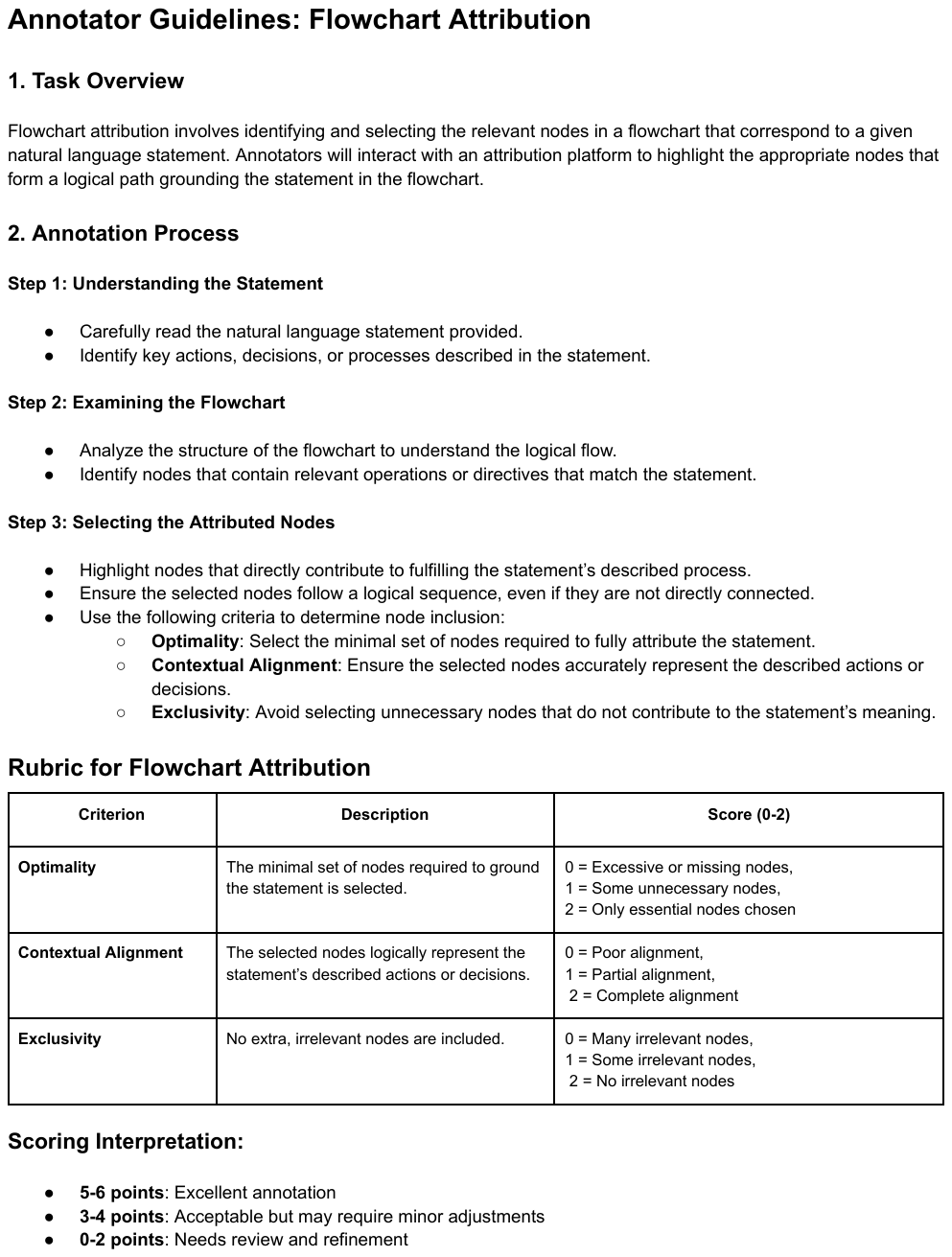}
    \caption{Summary of instructions given to human annotators.}
    \label{fig:annotator_guidelines_flowchart}
\end{figure*}

\begin{figure*}
    \centering
    \includegraphics[width=0.8\linewidth]{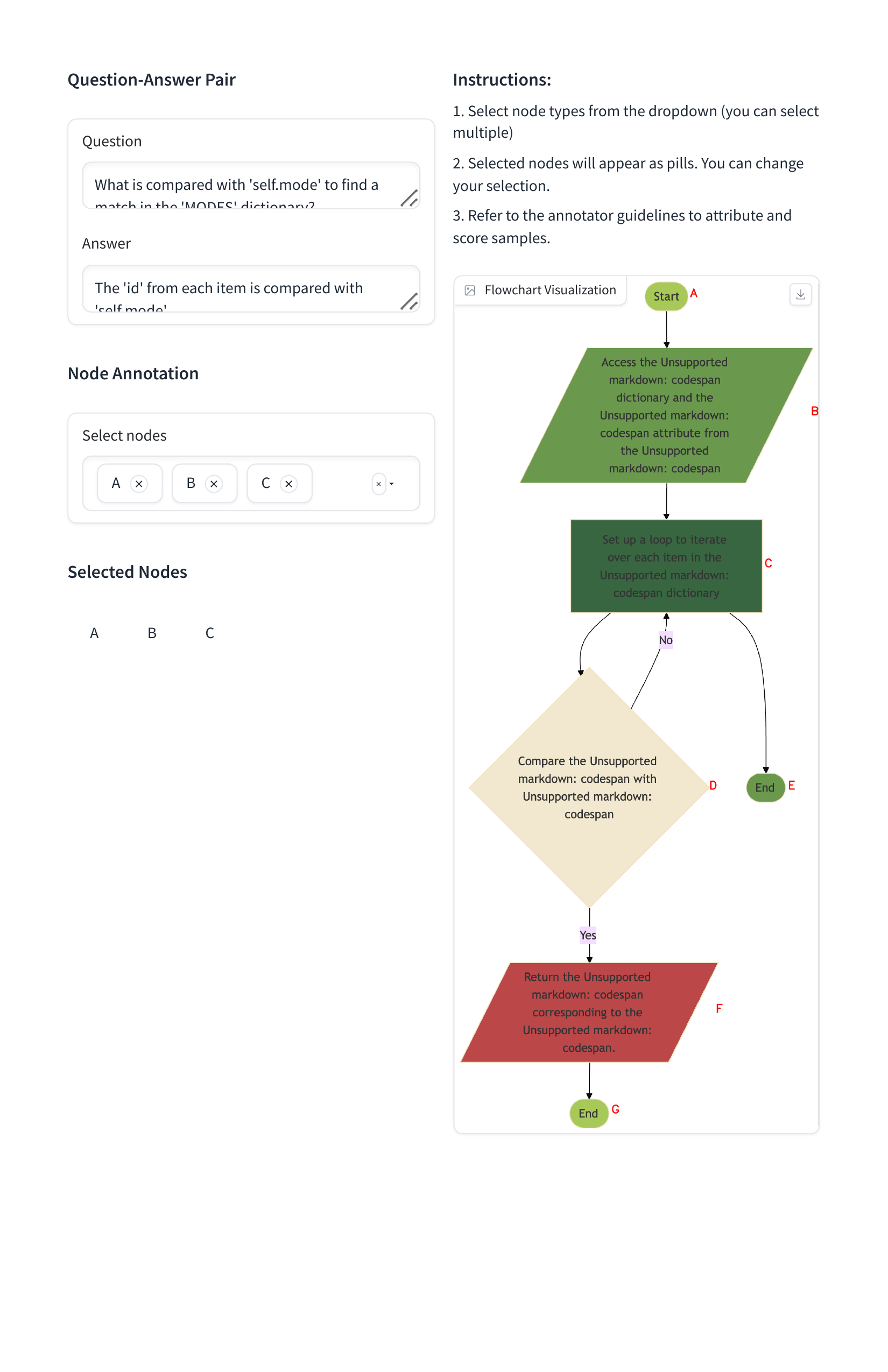}
    \caption{Human annotation platform for attribution annotation.}
    \label{fig:gradio_flowchart}
\end{figure*}

Fig \ref{fig:annotator_guidelines_flowchart} shows a summary of the annotator guidelines, and Fig \ref{fig:gradio_flowchart} shows the annotation platform used.

\label{sec:Appendix}

\section{Additional analysis on hand-constructed charts
}
\label{sec:hand}
We conducted a supplemntary case study to analyze \texttt{FlowPathAgent}'s generalization to real-world handwritten charts. 
\textbf{Data:} Given the lack of Question-Answer datasets for hand-drawn charts, we used the FC\_BScan \cite{bresler2016recognizing} dataset for hand-drawn flowchart component recognition. We randomly selected 50 samples from the test set, and used prompted GPT4o with example questions from \cite{flowvqa} to generate Question-Answer pairs. An annotator manually annotated ground-truth attributions for the selected samples. We used the train set to train FlowMask2Former for this domain.

\begin{figure}
    \centering
    \includegraphics[width=\linewidth]{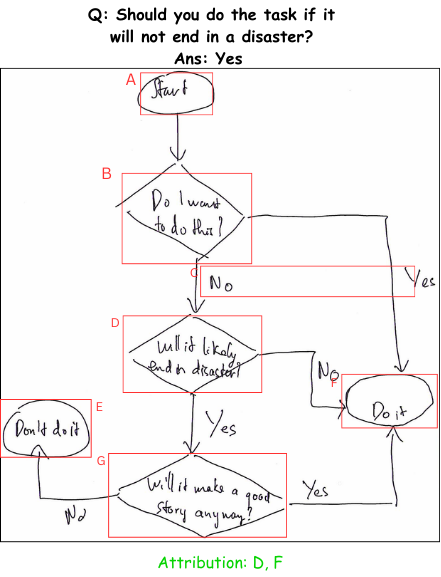}
    \caption{FlowPathAgent attributed D and F correctly. The blocks and labels represent FlowMask2Former annotations.}
    \label{fig:hand1}
\end{figure}

\begin{figure}
    \centering
    \includegraphics[width=\linewidth]{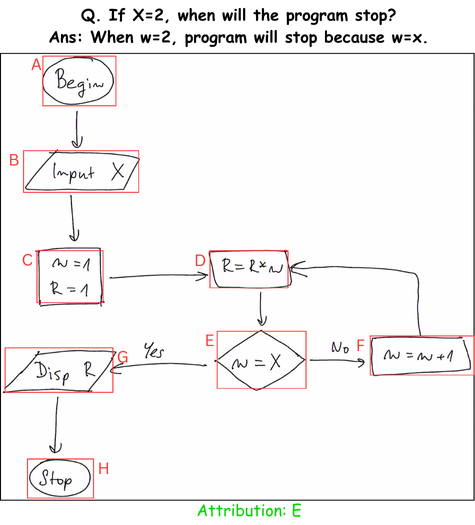}
    \caption{FlowPathAgent attributed E correctly. The blocks and labels represent FlowMask2Former annotations.}
    \label{fig:hand2}
\end{figure}

\begin{figure}
    \centering
    \includegraphics[width=\linewidth]{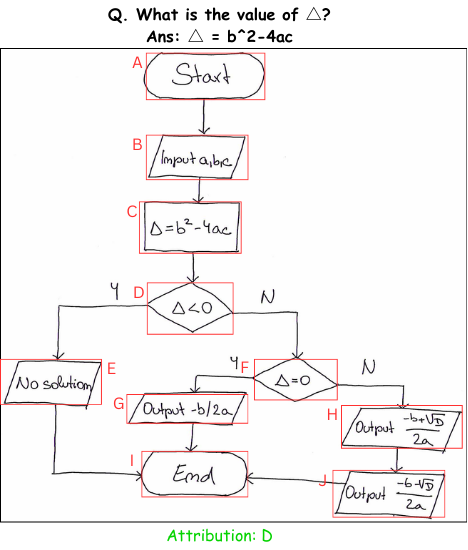}
    \caption{FlowPathAgent attributed D correctly. The blocks and labels represent FlowMask2Former annotations.}
    \label{fig:hand3}
\end{figure}

\begin{figure}
    \centering
    \includegraphics[width=0.7\linewidth]{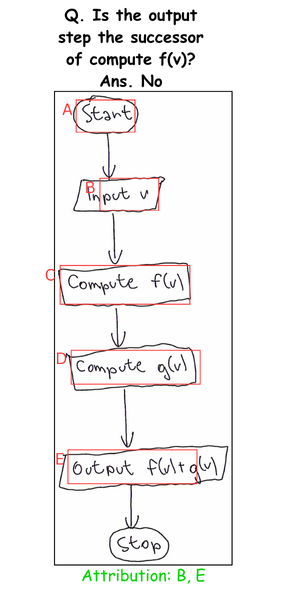}
    \caption{FlowPathAgent attributed B, and E correctly. The blocks and labels represent FlowMask2Former annotations.}
    \label{fig:hand4}
\end{figure}

\begin{figure}
    \centering
    \includegraphics[width=\linewidth]{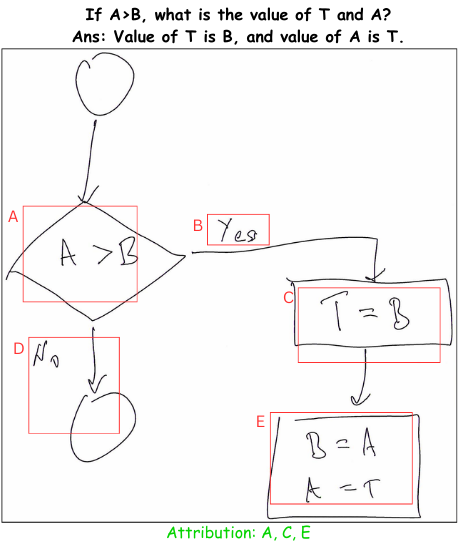}
    \caption{FlowPathAgent attributed A, B, C, E. The blocks and labels represent FlowMask2Former annotations.}
    \label{fig:hand5}
\end{figure}

Table~\ref{tab:case_study_handwritten} compares results from the chosen baselines.
Fig~\ref{fig:hand1}-\ref{fig:hand5} represent qualitative examples of \texttt{FlowPathAgent}'s performance. We observe, that due to the neuro-symbolic approach used by our agent, it is able to generalize across style variations and is is robust to errors in intermediate steps.

\begin{table}[t]
\centering
\resizebox{\columnwidth}{!}{%
\begin{tabular}{l|ccc}
\hline
\textbf{Baseline} & \textbf{Precision} & \textbf{Recall} & \textbf{F1} \\ \hline
Kosmos-2 \cite{peng2023kosmos} & 7.34 & 3.12 & 4.43 \\
LISA \cite{lai2024lisa} & 16.52 & 30.13 & 21.50 \\
SA2VA \cite{yuan2025sa2vamarryingsam2llava} & 22.12 & 5.13 & 8.25 \\
VisProg \cite{Gupta2022VisProg} & 0.00 & 0.00 & 0.00  \\
GPT4o Bounding Box & 32.37  & 32.44 & 32.41 \\
GPT4o SoM & 62.41 & 64.52 & 63.45 \\
\textbf{\texttt{FlowPathAgent}} & 65.11  & 68.25  & 66.64 \\ \hline
\end{tabular}%
}
\caption{Case Study: Performance comparison on Hand-written Charts
.}
\label{tab:case_study_handwritten}
\end{table}

\end{document}